\documentclass[twocolumn]{article}
\usepackage[a4paper, margin=0.7in]{geometry}
\usepackage{microtype}
\usepackage{graphicx}
\usepackage{subfigure}
\usepackage{booktabs} % for professional tables

\usepackage{hyperref}

\usepackage{amsmath}
\usepackage{amsthm}
\usepackage{amssymb}
\usepackage{mathtools}
\usepackage{xspace}
\usepackage{enumitem}
\setlist{leftmargin=*}
\setlist[itemize]{noitemsep}
\setlist[enumerate]{noitemsep}
\usepackage[dvipsnames]{xcolor}
\newcommand{\methodname}{SCRIB\xspace}
\usepackage{multirow}
\usepackage{capt-of}
\usepackage{wrapfig}

\usepackage{algorithm}
\usepackage{algorithmic}
\usepackage{array}
\usepackage{eqparbox}
\renewcommand{\algorithmiccomment}[1]{\hfill\eqparbox{COMMENT}{\{#1\}}}
\def\AlgoMain{

\begin{algorithm}[htb]
   \caption{Thresholds Finding for \methodname}
   \label{alg:main}
\textbf{Input}:\\
    $\mathbf{M} \in\mathbb{R}^{N\times K}$: model output on $\validationset$ sorted by column. $\mathbf{M}_{i,k}$ is the $i$-th smallest value in $\{m_k(x)\}_{x\in \validationset}$, for class $k$.
    $N$ denotes $|\validationset|$.
    \\
   $\hat{L}$ : $\mathbb{R}^K\mapsto \mathbb{R}$, empirical loss function on $\validationset$. \\
\textbf{Output}: \\
    $\mathbf{t} \in\mathbb{R}^K$: optimal thresholds for the set classifier $\mathbf{H}$.\\
\textbf{Algorithm}:
\begin{algorithmic}
    \STATE For $k\in [K]$, initialize $t_k$ randomly from $\mathbf{M}_{\cdot, k}$.
     \STATE Evaluate current loss $l\gets \hat{L}(\mathbf{t})$.
   \REPEAT
        \FOR{$k=1$ {\bfseries to} $K$}
        \STATE Fixing $t_{k'} \forall k'\neq k$, search $t'_k$ in $\mathbf{M}_{\cdot, k}$ to minimize $\hat{L}$ using QuickSearch (See Appendix for details)
        \STATE $l'_k \gets \hat{L}(\mathbf{t}_k')$ where $\mathbf{t}_k'\defeq (t_1, \ldots, t'_k, \ldots, t_K)$
        \ENDFOR
        \IF{$\min_{k\in [K]}{l'_k} < l$}
            \STATE Update $l \gets l'_k$ and $\mathbf{t} \gets \mathbf{t}_k'$
        \ENDIF
   \UNTIL {$l$ does not improve}
   \STATE {\bfseries return} $\mathbf{t}$
\end{algorithmic}
\end{algorithm}
}

\def \AlgoSearch{

\begin{algorithm}[htb]
   \caption{QuickSearch for threshold $t_d$ of class $d$}
   \label{alg:search_each_iter}
\textbf{Input}:\\
    $\mathbf{M} \in\mathbb{R}^{N\times K}$: sorted model output on $\validationset$. $\mathbf{M}_{i,k}$ is the $i$-th smallest value in $\{m_k(x)\}_{x\in \validationset}$, for class $k$.\\
    $\mathbf{I} \in\mathbb{R}^{N\times K}$: sorted indices such that $m_k(x_{\mathbf{I}_{j,k}}) = \mathbf{M}_{j,k}$\\
    $\mathbf{t}\in\mathbb{R}^K$: current thresholds\\
    $d\in[K]$: dimension to search \\
    $\{x_i, y_i\}_{i\in[N]}$: the data\\
\textbf{Output}: \\
    $t_d \in\mathbb{R}$: optimal threshold for class $d$ that minimizes the loss, fixing $\{t_k\}_{k\neq d}$\\
\textbf{Algorithm}:
\begin{algorithmic}
    \STATE Compute $sure_{k,1}$ and $err_{k,1}$ for $k\in[K]$, and $cnt_{i}$ for $i\in[N]$
    \STATE  $l^* \gets \infty$, $j^* \gets 0$ \COMMENT{ Initialize the optimum}
    \FOR {$j \gets 1, 2, \ldots, N - 1$ }
    \STATE $l\gets (1 - \frac{\sum_{k\in[K]}sure_{k,j}}{N})+ \sum_{k=1}^K\lambda_k (\frac{err_{k,j}}{sure_{k,j}}-r_k)_+^2$ \COMMENT{ Compute new loss $l$}
    \IF {$l< l^*$}
        \STATE $l^* \gets l$, $j^* \gets j$ \COMMENT{ Update the optimum if necessary}
    \ENDIF
    \STATE $i\gets I_{j+1, d}$ and $k\gets y_i$  \COMMENT{ map the $j$-th quantile back to the $i$-data point and its label}
    \IF[$cnt_i$ will become 1 (certain) after we change $t_d$ to $\mathbf{M}_{j,d} = m_d(x_i)$]{$cnt_i$ = 2} %(\tcp*[h]{comment})
    \STATE $err_k \gets err_k + \mathbf{1}\{d = k\} $ \COMMENT{[After $\mathbf{H}(x_i)$ becomes certain, it does not contain $y_i$]}
    \STATE $err_k \gets err_k + \mathbf{1}\{d \neq k \land  m_k(x_i)\leq t_k)\}$ \COMMENT{Same as above}
    \ELSIF[$cnt_i$ will become 0 (uncertain). The only label it contained must be $d$.]{$cnt_i$ = 1} %\COMMENT{$cnt_i$ will become 0 (uncertain)}
    \STATE $err_k \gets err_k - \mathbf{1}\{d \neq k\} $ \COMMENT{\# of errors can decrease as this prediction becomes uncertain.}
    \ENDIF
    \STATE $sure_k \gets sure_k + \mathbf{1}\{cnt_{i}=2\} - \mathbf{1}\{cnt_{i}=1\}$ \COMMENT{Update the \# of certain predictions}.
    \ENDFOR 
    \STATE {\bfseries return} $\mathbf{M}_{j^*, d}$
\end{algorithmic}
\end{algorithm} 
}
\def \FigExplainSetClassifier{
\begin{figure}[ht]
\vskip 0.2in
\centering
\includegraphics[width=1\columnwidth]{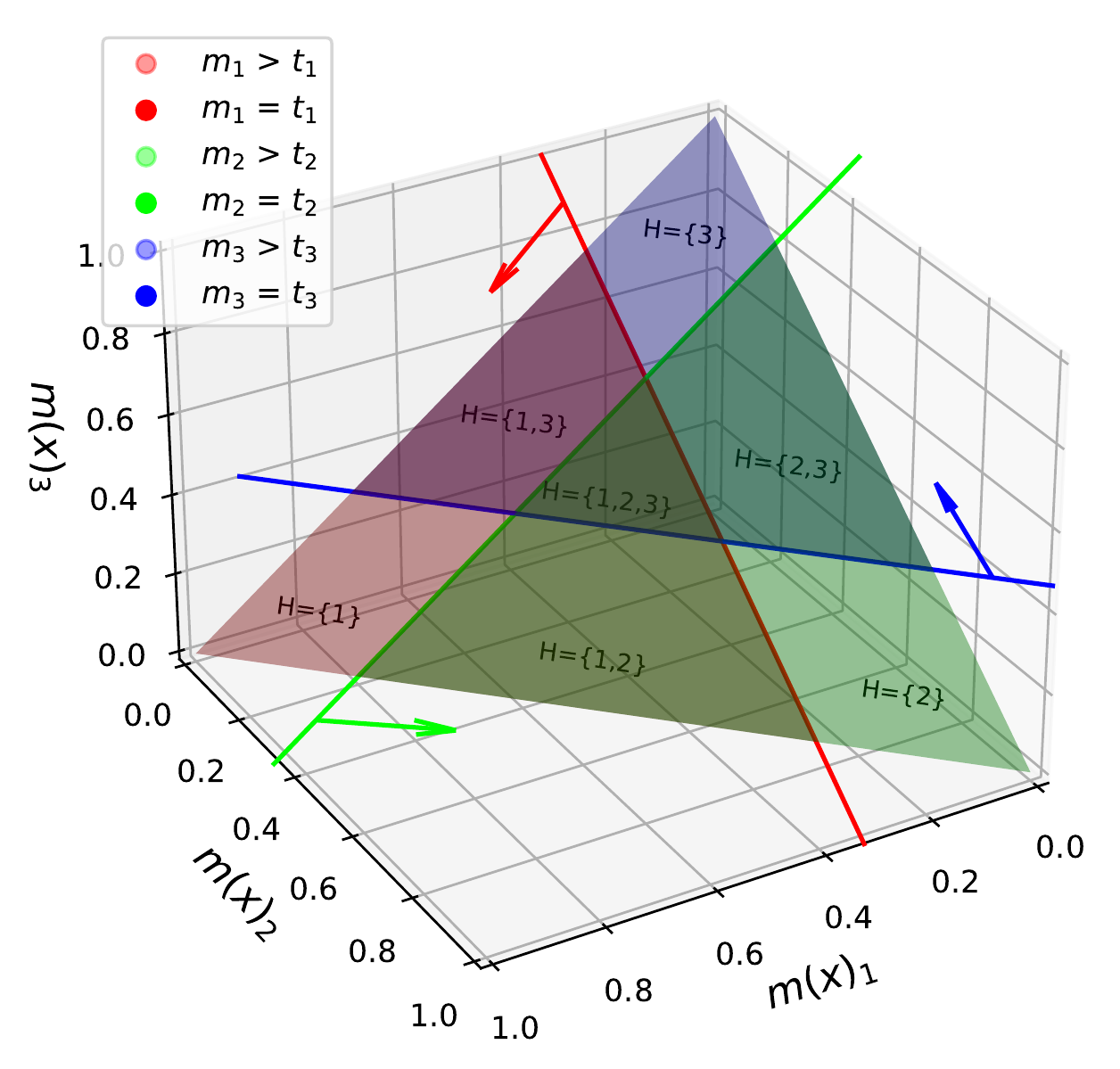}%
\caption{Here we show the set of possible predicted probabilities geometrically, with $K=3$ and $\mathbf{t} = [0.25, 0.2, 0.3]$.
Outputs $m(x)$ have been normalized to a valid probability distribution by Softmax in this case.
Intuitively, we draw $K$ hyperplanes in the form of $m_k(x) \geq \mathbf{t}_k$ to segment $\mathbb{R}^K$ into up to $2^K$ cells (some might be empty), and assign values to $\mathbf{H}(x)$ depending on which cell $m(x)$ falls in. 
In each cell, we show the value of the set classifier (E.g., $\mathbf{H}=\{1,2\}$).
}
\label{fig:explain_set}
\vskip -0.2in
\end{figure}
}
\def \NotationTable{
\begin{table}[ht]
\caption{Notations used in this paper}
\label{table:notation_table}
\begin{center}
\begin{small}
\resizebox{\columnwidth}{!}{
\begin{tabular}{cl}
\toprule
Symbol & Meaning \\
\midrule
$k$ & Class index \\
$[K]$ & The set $\{1,2,\ldots, K\}$\\
$\mathcal{X}, \mathcal{Y}$ & Data space and label space\\
$\mathbb{P}$ , $\mathbb{P}_k$ & Underlying data distribution (of class $k$)\\
$\mathbb{P}\{event\}$ & Probability of $event$ when data follows $\mathbb{P}$\\
$\mathbf{1}\{event\}$ & Indicator function of $event$\\
\hline
$\mathbf{H}$ & Set classifier: $\mathcal{X}\mapsto 2^\mathcal{Y}$\\
$A(\mathbf{H})$ & Ambiguity (Size- or Chance-) of $\mathbf{H}$\\
$r(\mathbf{H})$ , $r_k(\mathbf{H})$ & Risk of classifier $\mathbf{H}$ (of class $k$)\\
$r^*$, $r^*_k$ & Target risks (overall/for class $k$)\\
$m_k(x)$ & Base model prediction for $\mathbb{P}\{Y=k|X=x\}$\\
$\mathbf{t}$,$t_k$ & The threshold parameter for $\mathbf{H}$ (for class $k$)\\
$L(\mathbf{t})$ & Unconstrained loss given thresholds $\mathbf{t}$\\
$\hat{L},\hat{\mathbb{P}},\hat{A},\hat{r},\hat{r}_k$ & Empirical $L,\mathbb{P},A,r, r_k$ on $\validationset$\\
$\alpha_k(\mathbf{H})$ & Mis-coverage rate for $\mathbf{H}$ of class $k$\\
\bottomrule
\end{tabular}}
\end{small}
\end{center}
\end{table}
}

\def \TabDataClassCounts{
\begin{table}[ht]
\captionof{table}{
Total sample counts for each class of the validation dataset (excluding the training samples used to train the DL model).
}
\begin{center}
\begin{small}
\begin{tabular}{lccccr}
\toprule
Data \textbackslash Class & 0 & 1 & 2 & 3 & 4\\
\midrule
Xray      & 34 & 547 & 1,002 & 400 & N/A \\
ISRUC     & 4,907 & 2,857 & 7,255 & 4,476 & 2,985\\
SleepEDF  & 57,424 & 4,464 & 14,812 & 1,946 & 5,259\\
ECG       & 2,893 & 1,579 & 449 & 145 & N/A\\
\bottomrule
\end{tabular}
\label{table:exp:data:cnts}
\end{small}
\end{center}
\end{table}

}

\def \FigTabExpI{
\begin{figure}[ht]
    \centering
    \begin{subfigure}
        \centering 
        \includegraphics[width=0.23\textwidth]{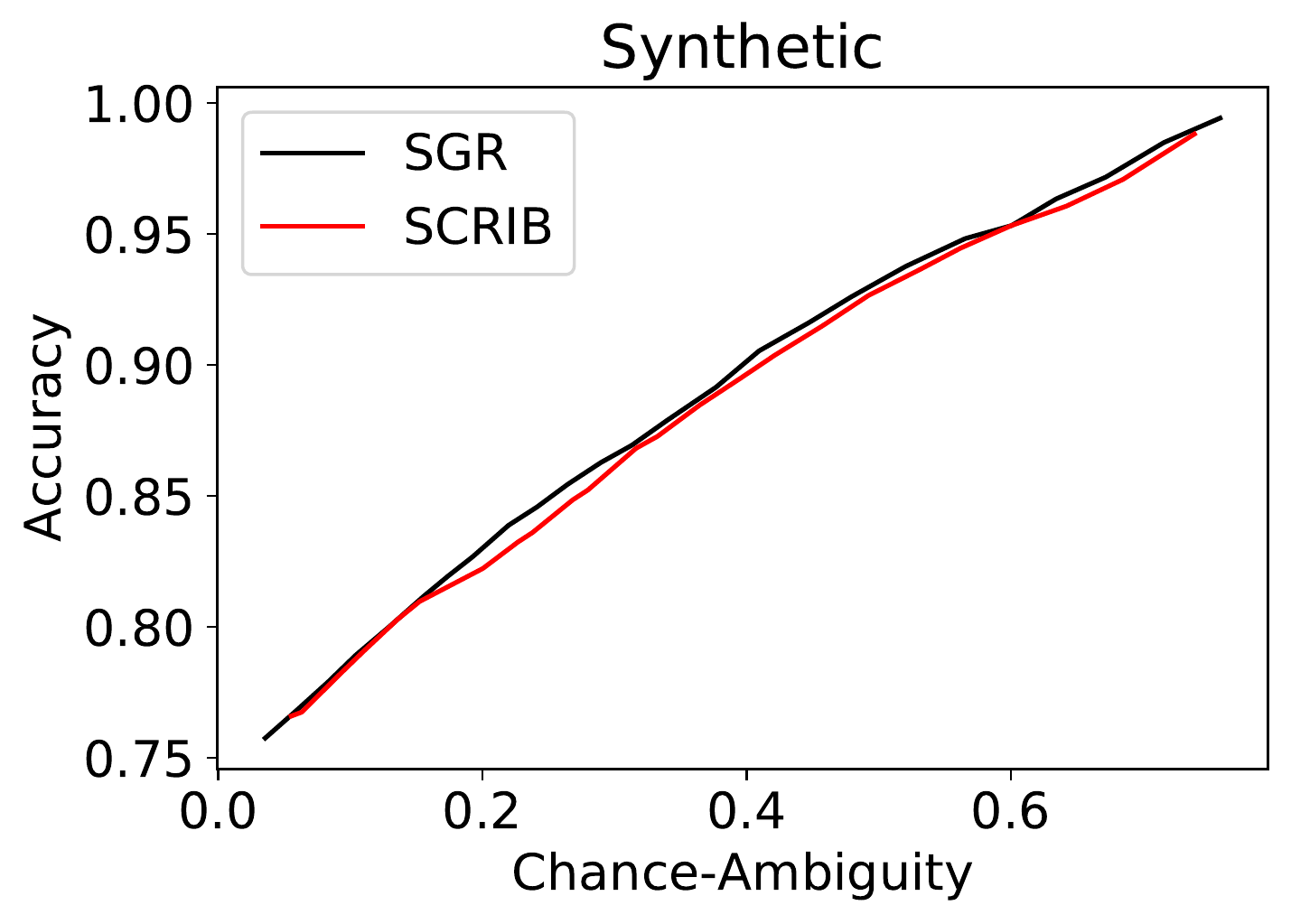} 
        \label{fig:exp1:syn2}
    \end{subfigure}
    \hfill
    \begin{subfigure}   
        \centering 
        \includegraphics[width=0.23\textwidth]{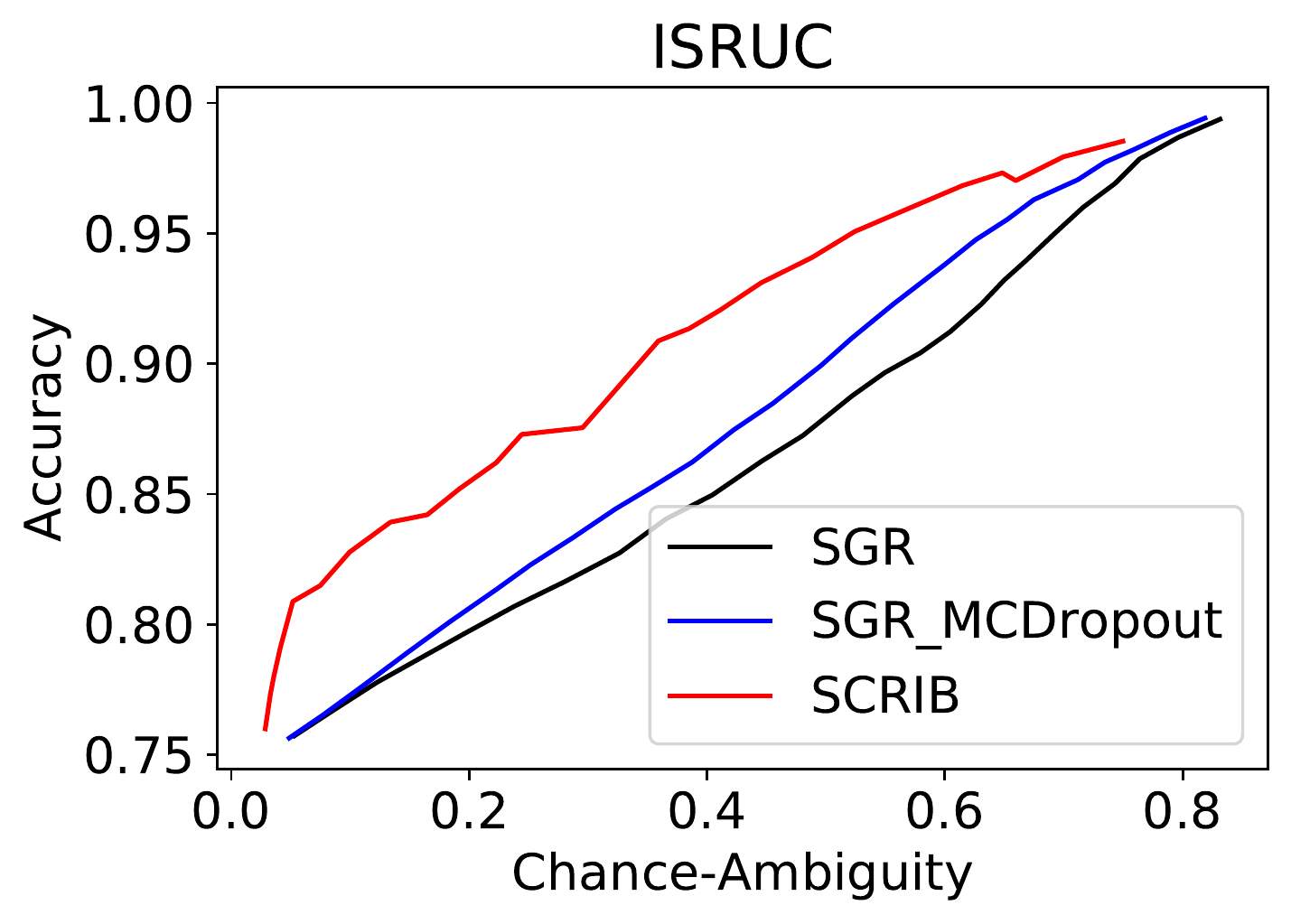} 
        \label{fig:exp1:isruc1}
    \end{subfigure}
    
\caption[ results for Experiment: Overall risk]
{
Accuracy-Ambiguity (reject rate) curve when we use different methods to control the overall risk. 
\methodname achieves similar or higher accuracy at the same level of ambiguity as SGR and its variant. 
}  

\label{fig:exp1}
\captionof{table}{Mean and standard deviation of AUC of the accuracy-ambiguity curve for different methods ($n=20$ experiments).
Statistically significant differences (at p=0.01) are bolded. 
AUC of \methodname is either comparable with SGR or sometimes even higher.
}
\begin{center}
\begin{small}
\begin{tabular}{l|ccc}
\toprule
AUC (1e-2) & SGR & SGR+Dropout & \methodname\\% & SGR & SGR+Dropout & \methodname \\
\midrule
Synthetic & \textbf{90.12}$\pm$0.43 & N/A             & 89.89$\pm$0.43\\ %p-val = 0
Xray              & 89.39$\pm$0.54 & N/A & 89.32$\pm$0.56\\ %p-val = 0.47
ISRUC                 & 87.55$\pm$0.71 & 88.60$\pm$0.56  & \textbf{90.77}$\pm$0.74\\ %p-val = 7e-15
SleepEDF              & 96.50$\pm$0.60 & 96.48$\pm$0.51 & 96.62$\pm$0.65\\ %p-val = 0.04
ECG              & 77.03$\pm$2.13 & N/A & \textbf{82.55}$\pm$0.67\\ %p-val =1.4e-9
\bottomrule
\end{tabular}
\label{table:exp_overall}
\end{small}
\end{center}
\end{figure}
}

\def \FigClassSpecArbitrary{

\begin{figure}[ht]
    \centering
    \begin{subfigure}
        \centering
        \includegraphics[width=0.23\textwidth]{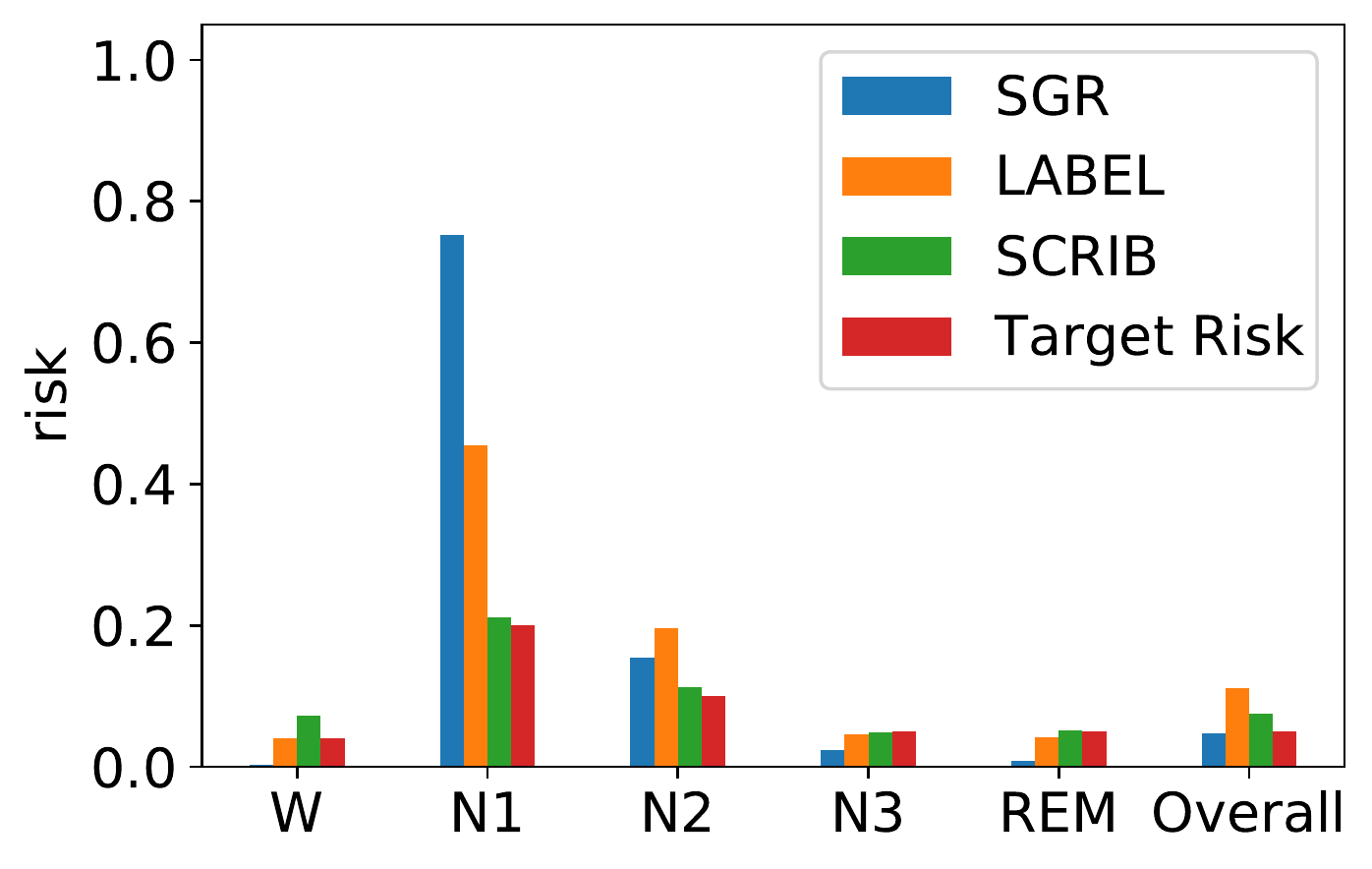}
        \label{fig:exp2:arb:isruc1}
    \end{subfigure}
    \hfill
    \begin{subfigure}
        \centering 
        \includegraphics[width=0.23\textwidth]{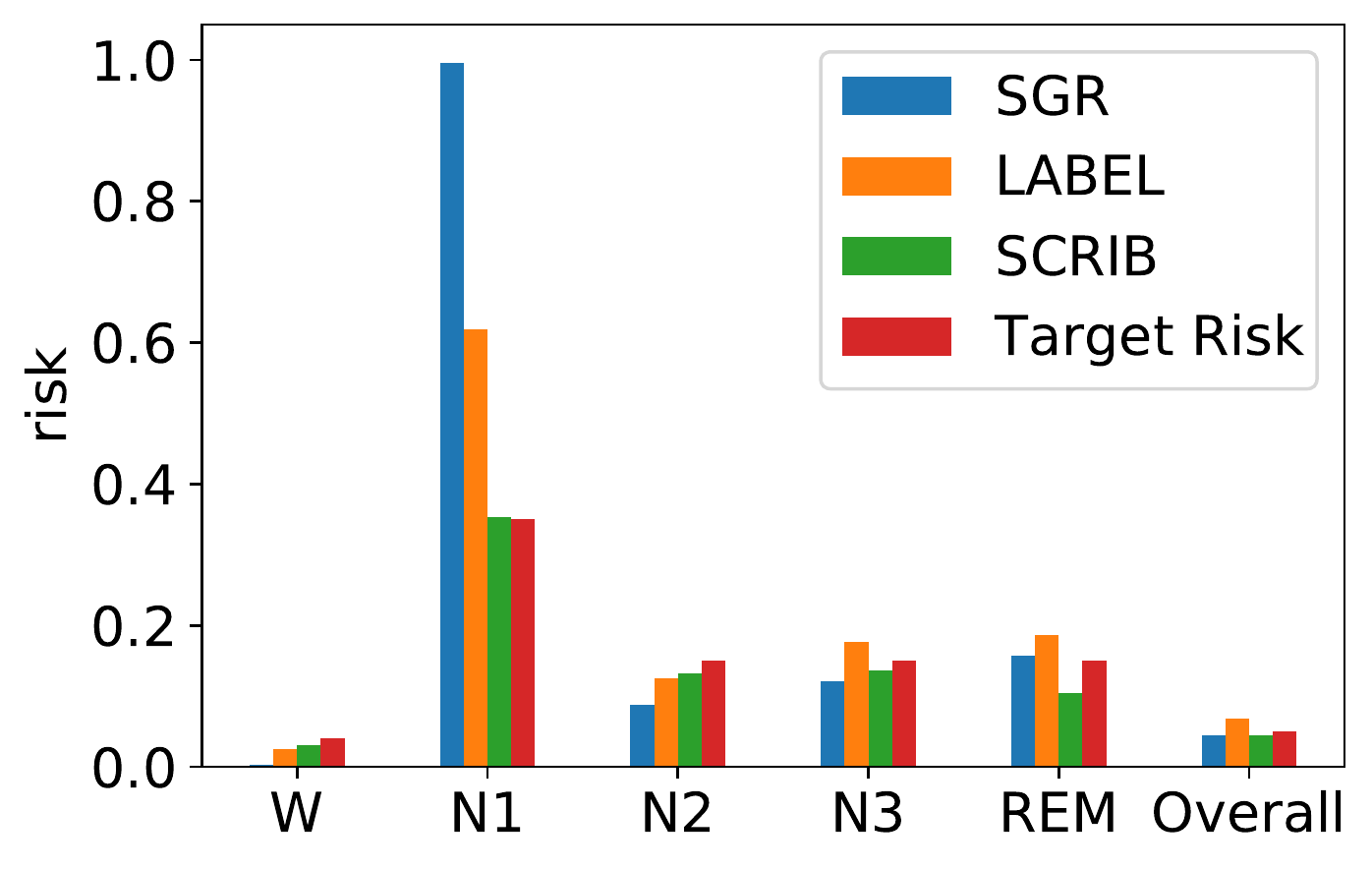}
        \label{fig:exp2:arb:edf1}
    \end{subfigure}
    
\caption[ Illustration of Risk Imbalance ]
{Class-specific risks for different methods on ISRUC (left) and Sleep-EDF (right).
Controlling only the overall risk (5\% in this case) leads to highly unbalanced class-specific risks (SGR).
It is possible to balance out risks without increasing the overall risk much with the targets we set (red columns). 
Here, the overall risk for SGR/LABEL/\methodname are 4.8\%/11\%/7.6\% for ISRUC, and 4.5\%/6.8\%/4.4\% for Sleep-EDF. 
However, \methodname achieves much lower risk on N1 the hard class than SGR and LABEL.
} 
\label{fig:exp2:arb}
\end{figure}
}

\def \FigTabExpII{
\begin{figure}[ht]
    \centering
    \begin{subfigure}
        \centering
        \includegraphics[width=0.23\textwidth]{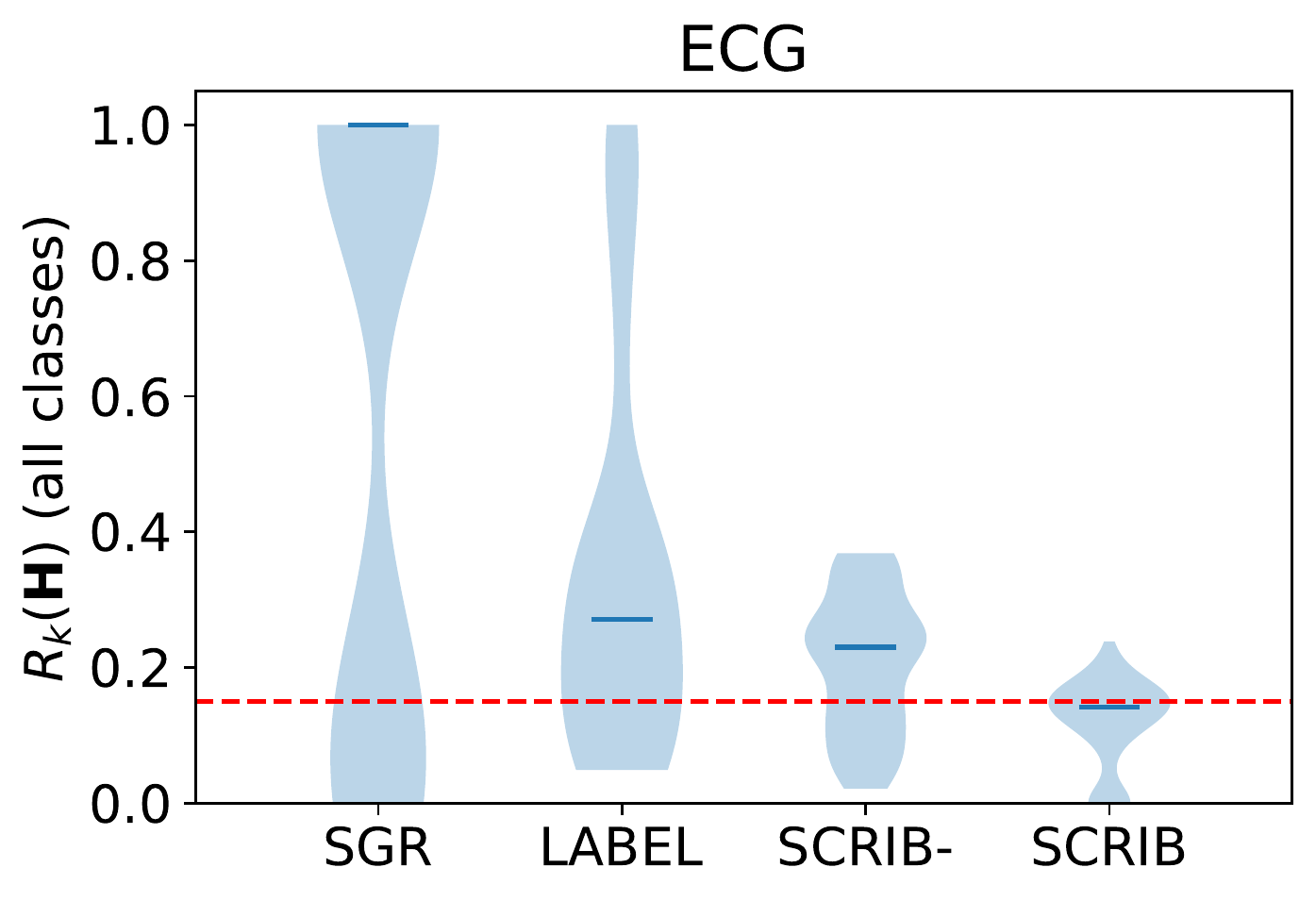}
        \label{fig:exp2:isruc1}
    \end{subfigure}
    \hfill
    \begin{subfigure}
        \centering 
        \includegraphics[width=0.23\textwidth]{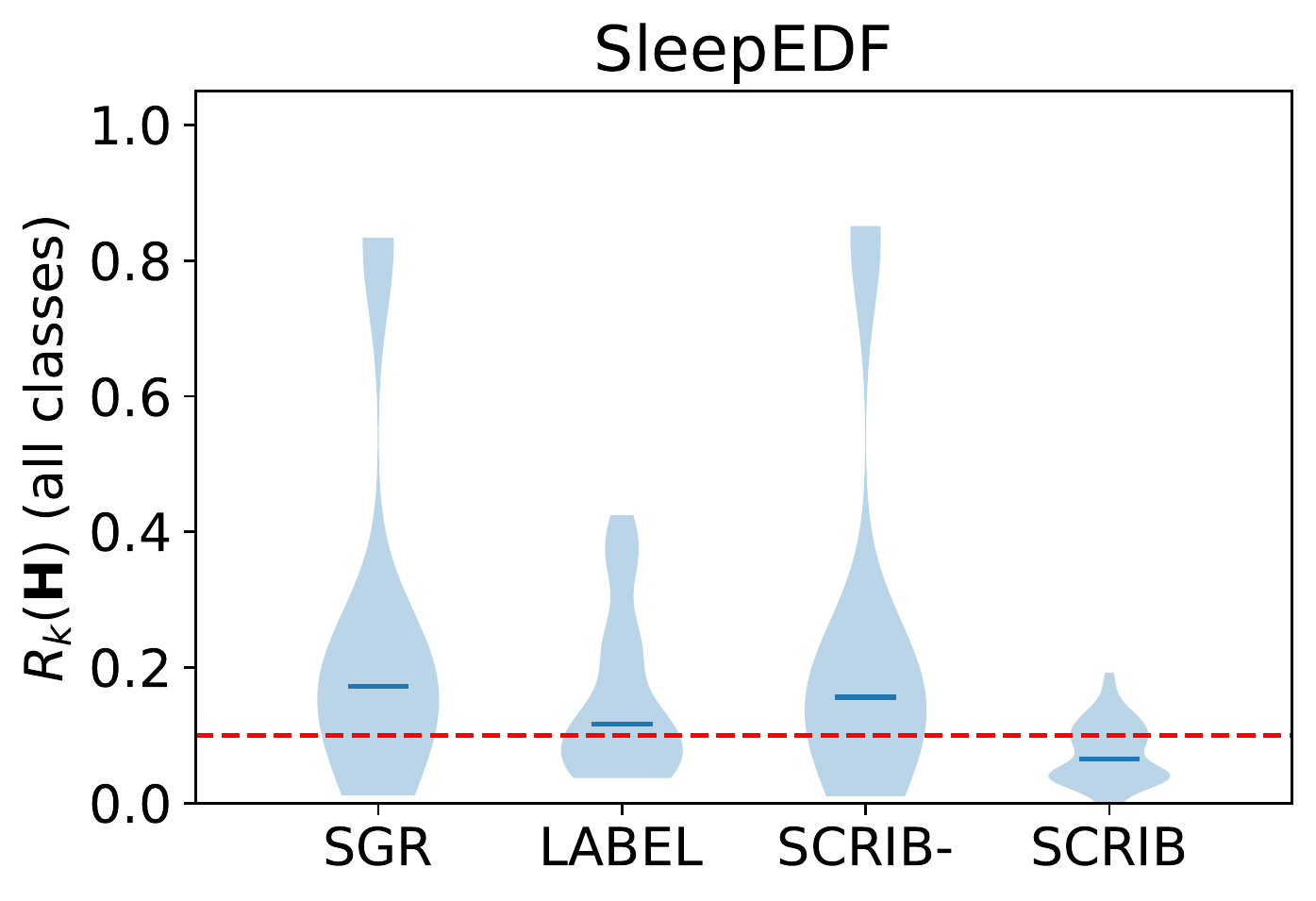}
        \label{fig:exp2:edf1}
    \end{subfigure}
    
\caption[ results for Experiment: Class-specific risks ]
{
Distribution (violin plots) of class-specific risks for different methods with marked medians.
Realized class-specific risks of \methodname are concentrated around/below target (red dashed lines), unlike other methods. Especially SGR has to reject almost all examples from the difficult classes (i.e., class-specific risk $R_k(H)$ is large).
} 
\label{fig:exp2}
\captionof{table}{Average class specific excess risk ($(\Delta r_k)_+$ in percentage) for each methods. 
The p-values for the two-sample mean t-test between \methodname and the best baseline are reported in parenthesis.
\methodname directly controls the risk and has lower deviations from targets than all baselines. Note that \methodname- with a global threshold cannot control the class-specific risk, which confirms the need for class-specific thresholds as in \methodname. }
\begin{center}
\begin{small}
\begin{tabular}{lcccr}
\toprule
$(\Delta r_k)_+$ (\%) & SGR & LABEL & \methodname- & \methodname\\% & SGR & SGR+Dropout & \methodname \\
\midrule
Xray        & 5.67 & 6.43 & 13.69 & 3.71 (0.34)\\ %(n=80)
ISRUC       & 8.60 & 4.23 & 8.79 & \textbf{1.78} (0.01)\\ %(n=100)
SleepEDF    & 16.90 & 7.21 & 16.32 & \textbf{0.89} (3e-8)\\ %(n=100)
ECG         & 46.23 & 21.58 & 7.71 & \textbf{0.90} (6e-12)\\ %(N=80)
\bottomrule
\end{tabular}
\label{table:exp_classSpec}
\end{small}
\end{center}
\end{figure}

}

\def \FigUserViolin{

\begin{figure}[ht]
    \centering
    \includegraphics[width=0.3\textwidth]{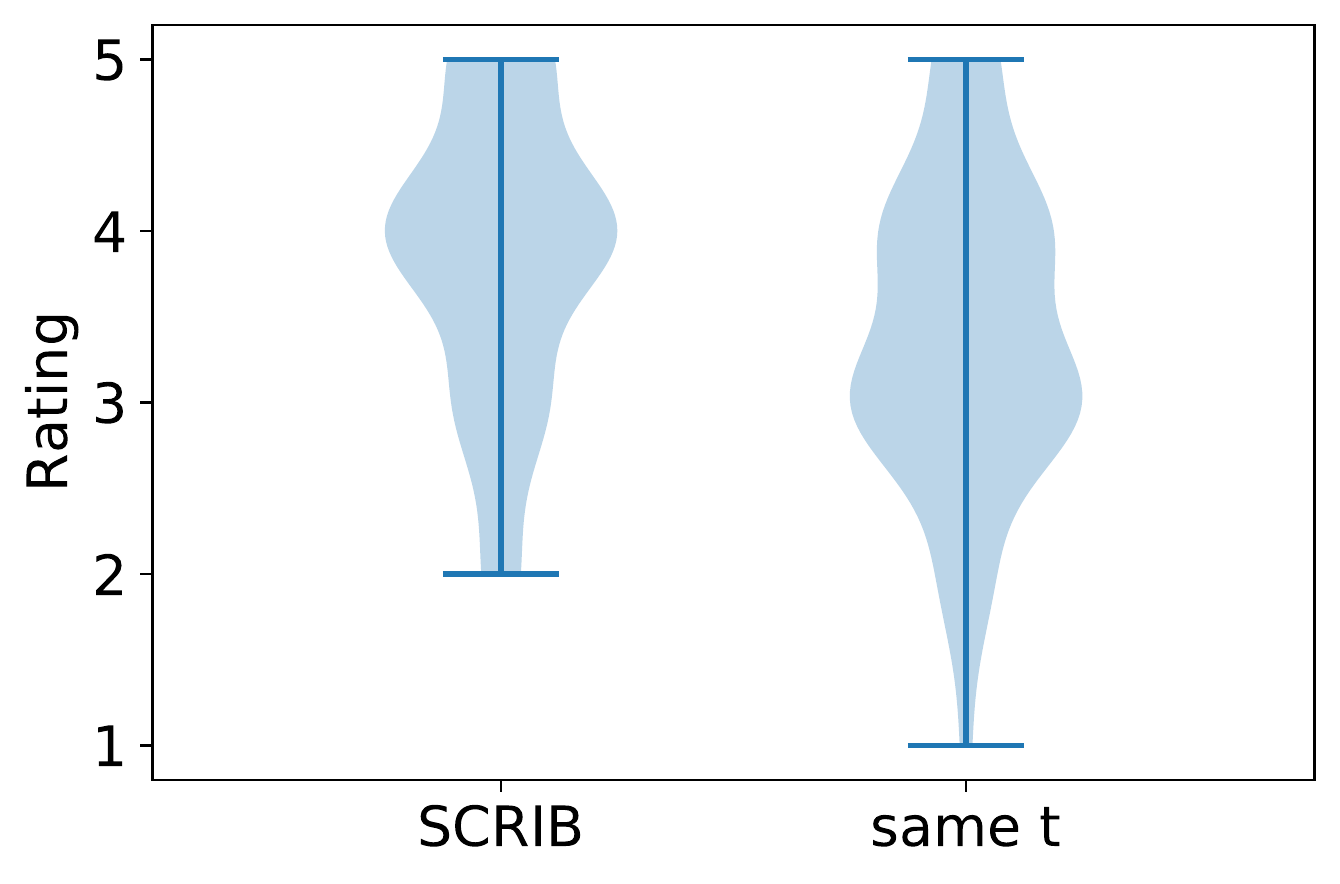}
\caption[ Scoring Distribution ]
{
Distribution of ratings given by doctor. 
\methodname (left) has higher ratings than using the same threshold for all classes (right). 
} 
\label{fig:exp_user}
\end{figure}
}

\def \AmbiguityCorrelation{
\begin{table}[ht]
\caption{Pearson and Spearman Correlation on the test set between Chance- and Size-Ambiguity among 1000 randomly generated $\mathbf{H}$ on the validation set. 
Numbers are in percentages.
In all real data-sets, the two ambiguities are highly positively correlated.
Means and standard deviations are reported for 20 experiments.}
\label{table:exp_ext:ambireal}
\vskip 0.15in
\begin{center}
\begin{small}
\begin{tabular}{l|cc|r}
\toprule
Dataset & Pearson & Spearman \\
\midrule
Xray & 90.07 $\pm$ 0.73 & 90.74 $\pm$ 0.92 \\
ISRUC & 85.46 $\pm$ 0.94 & 85.56 $\pm$ 1.17  \\
SleepEDF & 82.82 $\pm$ 0.98 & 81.08 $\pm$ 1.23  \\
ECG & 90.28 $\pm$ 0.58 & 91.39 $\pm$ 0.78  \\
\bottomrule
\end{tabular}
\end{small}
\end{center}
\vskip -0.1in
\end{table}

}

\def \TabExpExtendedIRMSE{
\begin{table}[ht]
\captionof{table}{
For the overall risk control experiment, we compute the RMSE between the realized risks and target risks in 20 experiments for each dataset and report the mean and standard deviation.
All methods tend to deviate from the original overall risk target by about 1-2\%, except for ECG, where the discrepancy is usually bigger.
SGR+Dropout does not apply to three datasets for the same reasons mentioned in the main text - we either do not have access to the model or there is no curve due to intrinsic issues with MCDropout.
}
\begin{center}
\begin{small}
\begin{tabular}{lccr}
\toprule
\hline 
$RMSE$ (\%) & SGR & SGR+Dropout & \methodname \\
\midrule
Synthetic & 0.72$\pm$0.40 & N/A & 0.71$\pm$0.20\\
Xray      & 2.43$\pm$0.88 & N/A & 2.31$\pm$0.98\\
ISRUC     & 1.56$\pm$0.99 & 1.78$\pm$1.14 & 1.74$\pm$0.95\\
SleepEDF  & 2.59$\pm$1.50 & 2.38$\pm$1.00 & 2.54$\pm$1.53\\
ECG       & 7.52$\pm$2.27 & N/A & 4.9$\pm$2.15\\
\hline
\bottomrule
\end{tabular}
\label{table:exp_overall_ext:riskRMSE}
\end{small}
\end{center}
\end{table}

}

\def \TabExpExtendedII{
\begin{table}[ht]
\captionof{table}{Average class specific excess risk ($(\Delta r_k)_+$ in percentage) for each methods, split by subjects.
The p-values for the two-sample mean t-test between \methodname and the best baseline are reported in parenthesis.
\methodname still greatly reduces the excess class-specific risks, and the difference is usually significant.
}
\begin{center}
\begin{small}
\begin{tabular}{lcccr}
\toprule
$(\Delta r_k)_+$ (\%) & SGR & LABEL & \methodname- & \methodname\\% & SGR & SGR+Dropout & \methodname \\
\midrule
ISRUC       & 8.55 & 5.48 & 9.06 & \textbf{2.54} (8e-4)\\ %(n=100)
SleepEDF    & 18.18 & 9.93 & 17.43 & 7.45 (0.21)\\ %(n=100)
ECG         & 42.98 & 21.62 & 7.71 & \textbf{2.55} (1e-6)\\ %(N=80)
\bottomrule
\end{tabular}
\label{table:exp_classSpec_ext:split_subjects}
\end{small}
\end{center}
\end{table}

}

\def \TabExpExtendedIIAmbiguity{
\begin{table}[ht]
\captionof{table}{Ambiguities of different methods. 
\methodname has higher but comparable ambiguities compared with baselines. 
This is, however, expected, because the underlying classifier is the same and trade-off is unavoidable. 
For example, choosing a global threshold for SleepEDF can reduces the Chance-Ambiguity greatly at 10\% risk level, because base classifier's risk without rejection is already close to 10\%. 
However, as we saw in main text, this incurs high risk for N1 class, so even mildly controlling its risk (using LABEL) will increase the ambiguity significantly. 
This is why we include the Overall Risk Experiment to show that with the same target, \methodname's ambiguity is very comparable. 
}
\begin{center}
\begin{small}
\begin{tabular}{l|cccc|cccr}
\toprule
\hline 
 &  \multicolumn{4}{c|}{Chance-Ambiguity}  &  \multicolumn{4}{c}{Size-Ambiguity} \\
 & SGR & LABEL & \methodname- & \methodname & SGR & LABEL & \methodname- & \methodname \\
\midrule
Xray   & 0.56 & 0.38 & 0.78 & 0.64  %(8e-7)
& 2.67 & 1.38 & 2.83 & 2.31 \\ %(n=100)
ISRUC   & 0.43 & 0.42 & 0.30 & 0.71  %(8e-7)
& 2.72 & 1.45 & 1.33 & 1.81 \\ %(n=100)
SleepEDF & 0.01 & 0.20 & 0.03 & 0.34 %(3e-8)     
& 1.05 & 1.25 & 1.05 & 1.68 \\ %(n=100)
ECG     &  0.97 & 0.66 & 0.57 & 0.80 %(9e-6)    
& 3.91 & 1.67 & 2.70 & 3.39 \\ %(N=80)
\hline
\bottomrule
\end{tabular}
\label{table:exp_classSpec_ext:ambiguities}
\end{small}
\end{center}
\end{table}

}

\def \TabOptimizationCompareNew{
\begin{table}[ht]
\footnotesize
\captionof{table}{
Final loss value and total search time of different methods, all with  $\geq 2$ significant figures. 
Mean and standard deviation computed with 40 experiments.
Our optimization method consistently performs better than alternatives and uses comparable search time.
}
\begin{center}
\begin{small}
\begin{tabular}{lc|cccccr}
\toprule
\hline 
 & & Bayesian & TNC & Powell & L-BFGS-B & Ours- & Ours\\%& Bayesian & TNC & Powell & L-BFGS-B & Ours \\
\midrule
\multirow{4}{*}{Loss}
& Xray    & 118$\pm$93 & 338$\pm$197 & 5.0$\pm$4.2 & 339$\pm$197 & 2.9$\pm$2.7 & \textbf{0.70} $\pm$0.012\\
& ISRUC   & 544$\pm$185 & 1356$\pm$736 & 3.1$\pm$6.7 & 1356$\pm$736 & 2.0$\pm$2.9 & \textbf{0.73} $\pm$0.33 \\
& SleepEDF& 1189$\pm$290& 2116$\pm$2144 & 798$\pm$3384 & 2117$\pm$2144 & 0.43$\pm$0.20 & \textbf{0.35} $\pm$0.010\\
& ECG     & 342$\pm$56  & 895$\pm$211 & 41$\pm$34 & 895$\pm$211 & 2.1$\pm$3.0 & \textbf{0.78} $\pm$0.013\\
\hline
\multirow{4}{*}{\vtop{\hbox{Time}\hbox{(second)}}}
& Xray    & 31$\pm$6.7  & 0.039$\pm$0.007   & 1.51$\pm$0.15 & 0.019$\pm$0.0051 & 0.10$\pm$0.012 & 1.43$\pm$0.21\\
& ISRUC   & 48.$\pm$8.9  & 0.22$\pm$0.027    & 14$\pm$1.5 & 0.097$\pm$0.010 & 3.8$\pm$0.68 & 10.$\pm$1.3\\
& SleepEDF& 46$\pm$7.9  & 14$\pm$5.3     & 96$\pm$9.9 & 1.2$\pm$0.40 & 6.3$\pm$1.5 & 35$\pm$4.9\\
& ECG     & 30.$\pm$9.1  & 0.063$\pm$0.0084 & 2.9$\pm$0.40 & 0.031$\pm$0.0075 & 0.57$\pm$0.086 & 2.3$\pm$0.36\\
\hline
\bottomrule
\end{tabular}
\label{table:exp_opt:loss_and_time_new}
\end{small}
\end{center}
\end{table}
}

\def \FigOptimizationCompare{

\begin{figure}[ht]
    \centering
    \begin{subfigure}
        \centering
        \includegraphics[width=0.45\textwidth]{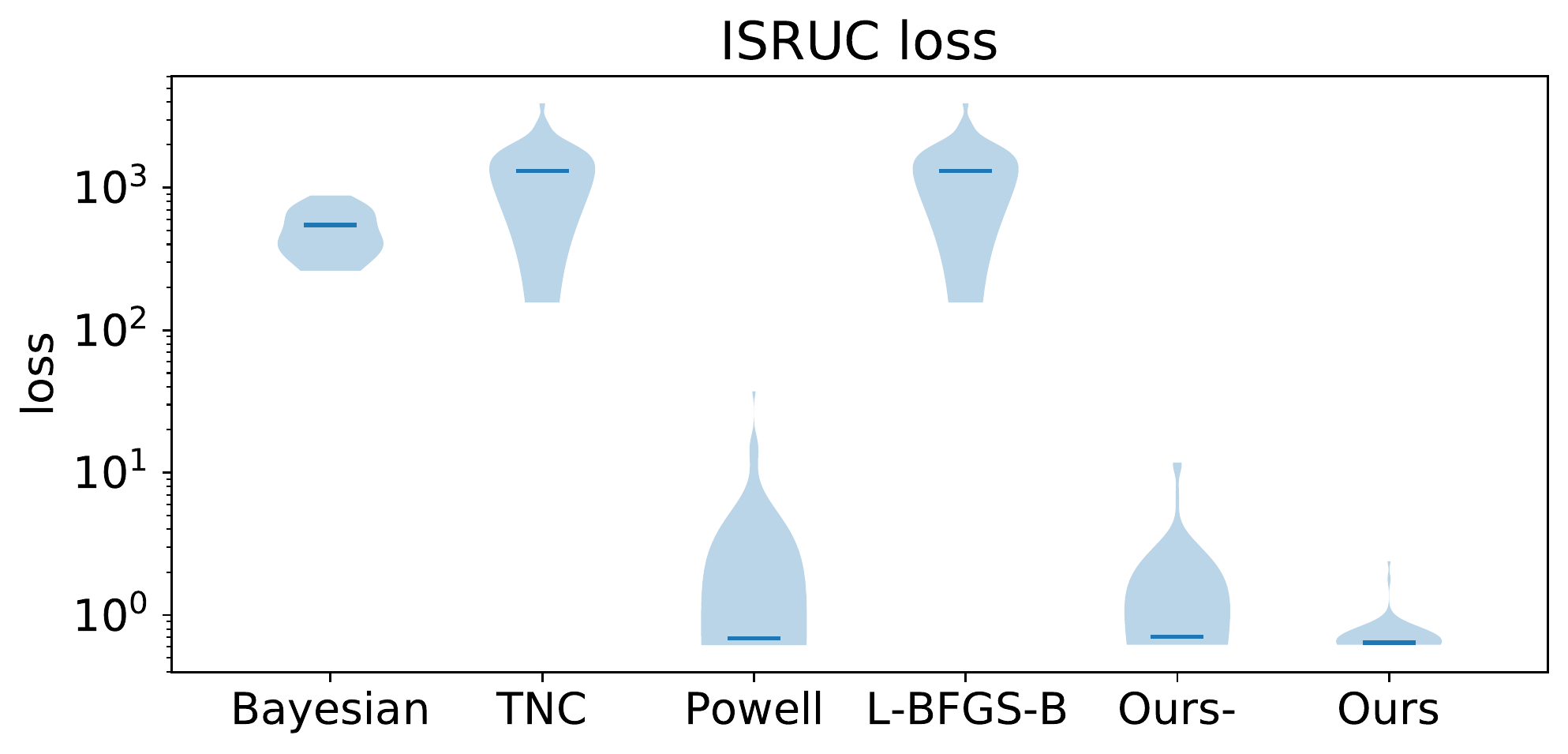}
        \label{fig:exp_opt:isruc}
    \end{subfigure}
    \hfill
    \begin{subfigure}
        \centering 
        \includegraphics[width=0.45\textwidth]{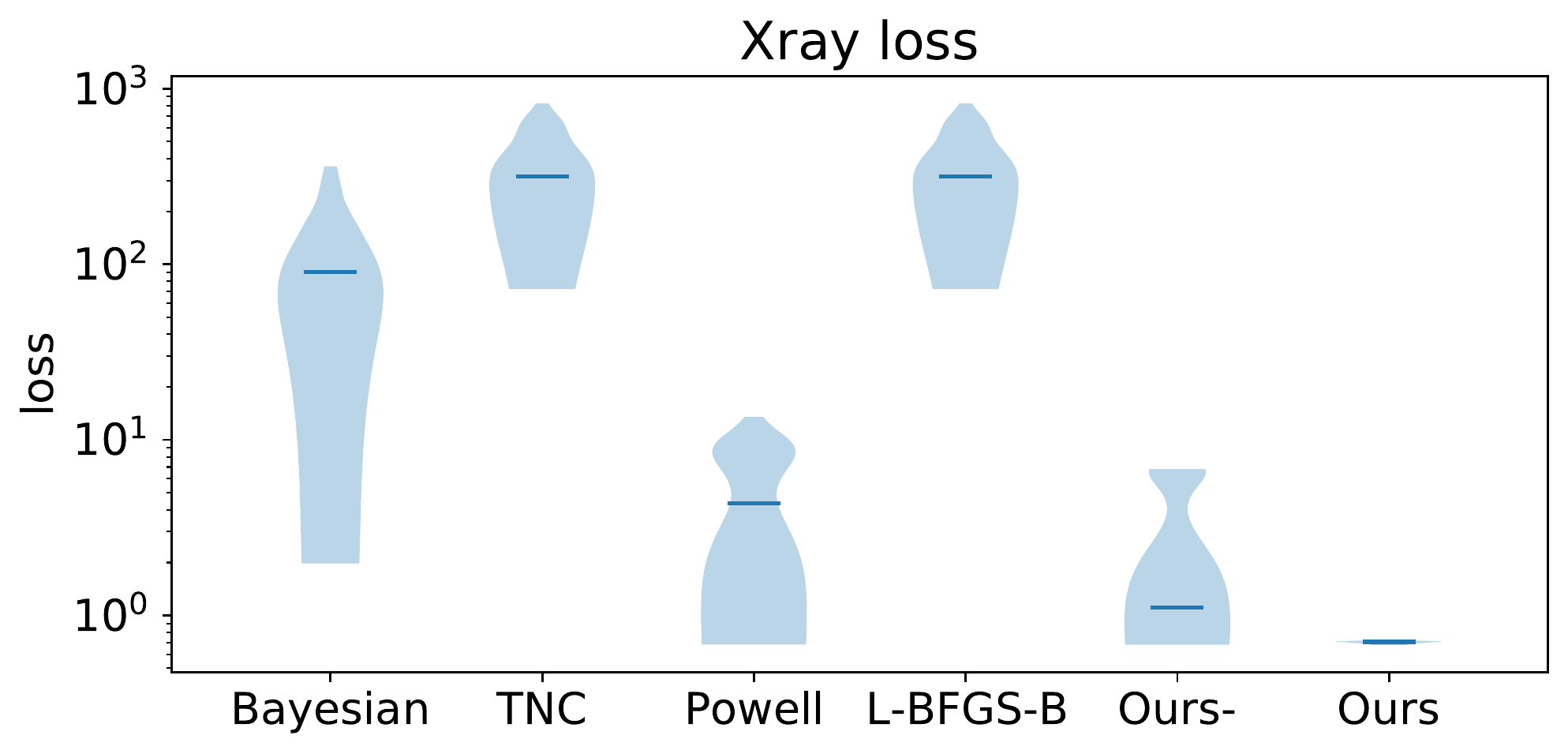}
        \label{fig:exp_opt:xray}
    \end{subfigure}
    \\
    \begin{subfigure}
        \centering
        \includegraphics[width=0.45\textwidth]{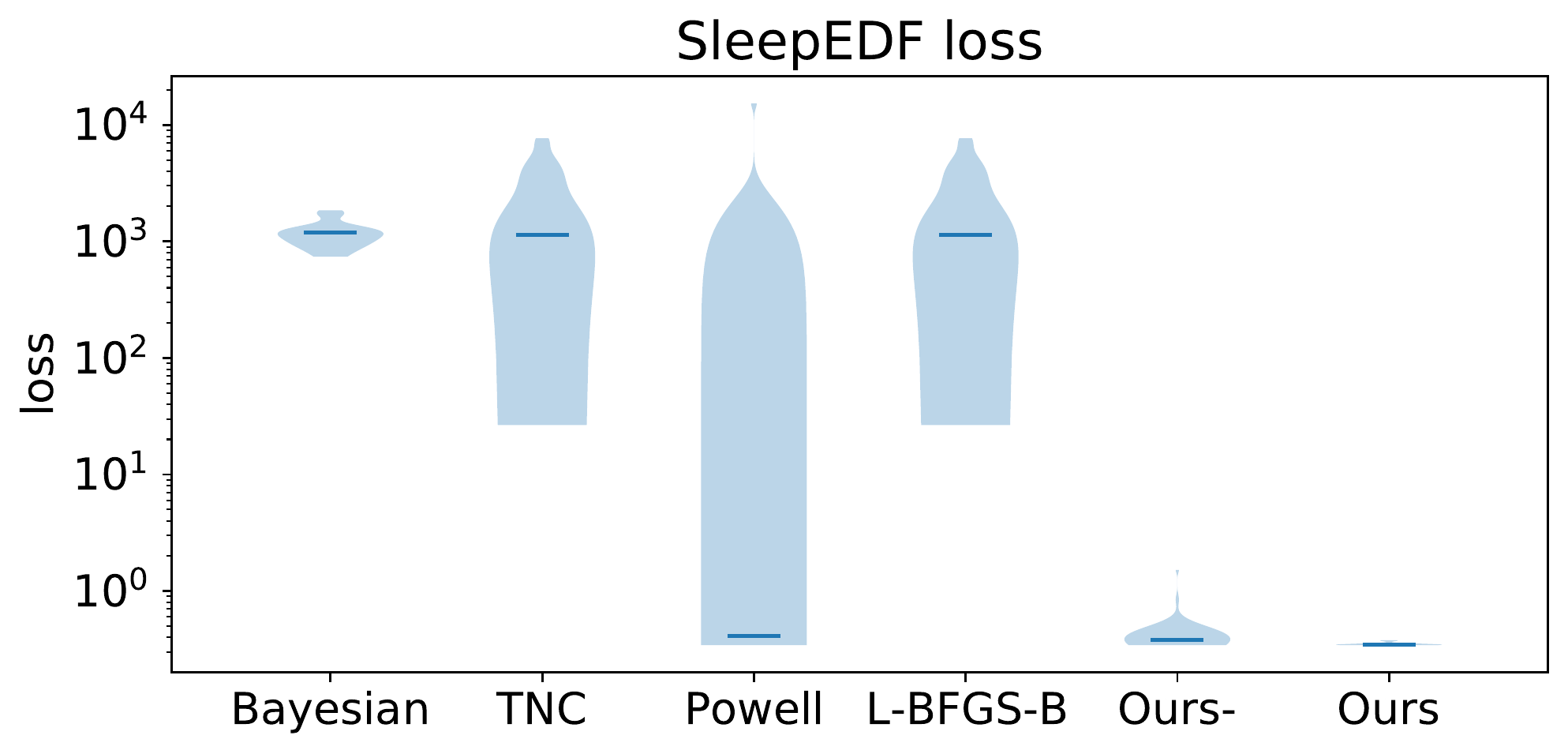}
        \label{fig:exp_opt:edf}
    \end{subfigure}
    \hfill
    \begin{subfigure}
        \centering 
        \includegraphics[width=0.45\textwidth]{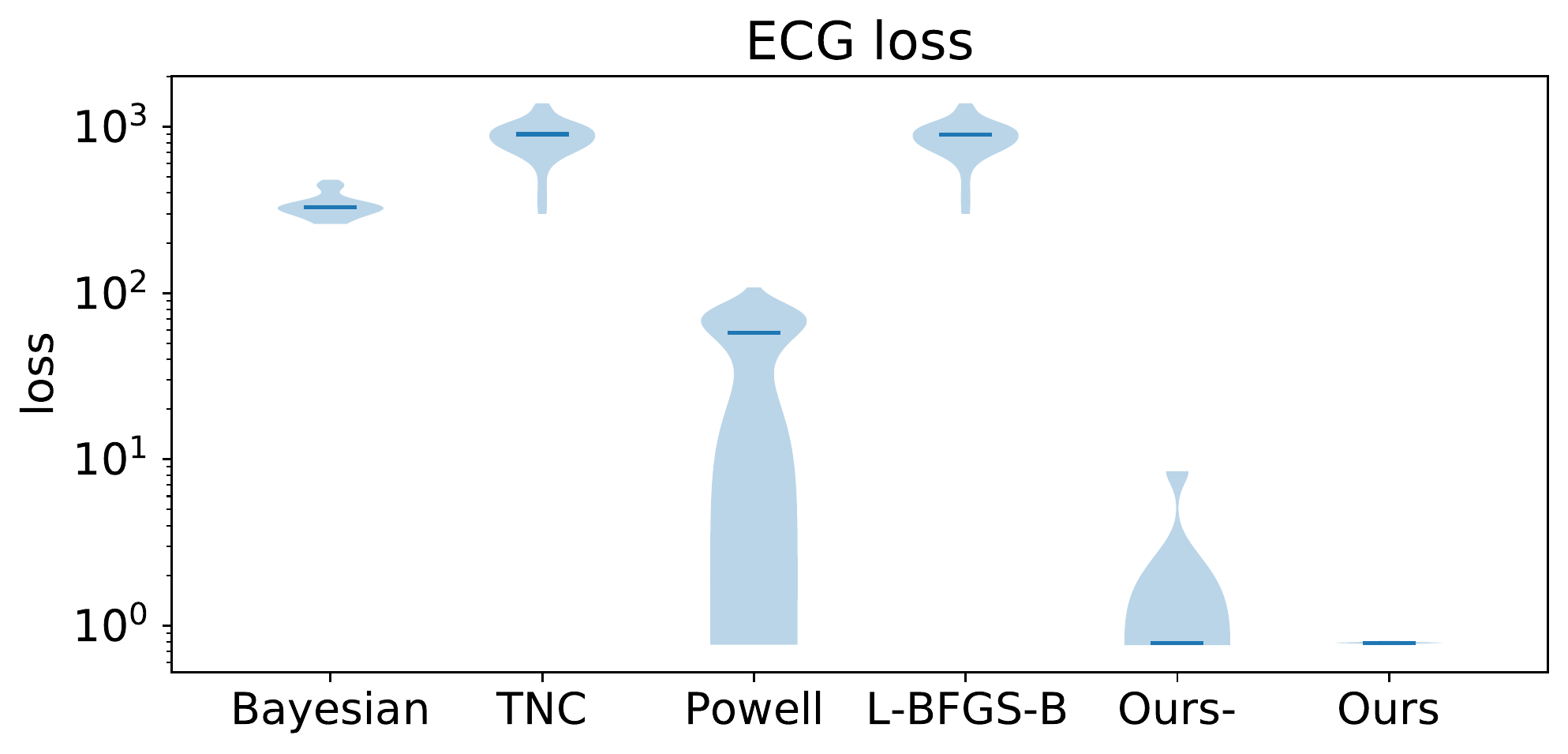}
        \label{fig:exp_opt:ecg}
    \end{subfigure}
\caption[ Illustration of optima distribution ]
{
Distribution of final loss in 40 experiments, with medians marked in dark blue. 
The axes are in log-scale.
Our optimization method achieves lower loss than all alternative methods.
The closest alternative is Powell, but it has much larger variances.
} 
\label{fig:exp_opt}
\end{figure}
}

\newcommand{\defeq}{\vcentcolon=}

\newtheorem{theorem}{Theorem}[section]

\newtheorem{lemma}[theorem]{Lemma}
\newtheorem{definition}{Definition}
\newtheorem{problem}{Problem}
\newcommand{\validationset}{\mathcal{S}_{\text{valid}}} 
\newcommand{\testset}{\mathcal{S}_{\text{test}}} 
\newcommand{\trainingset}{\mathcal{S}_{\text{train}}}

\def \uiuccs {University of Illinois at Urbana–Champaign}
\def \iqvia {Analytics Center of Excellence, IQVIA}
\def \mgh {Massachusetts General Hospital}
\author{
Zhen Lin$^{1}$,
Cao Xiao$^{2}$,
Lucas Glass$^{2}$,
M. Brandon Westover$^{3}$,
Jimeng Sun$^{1}$\thanks{Contact e-mail: \href{mailto:jimeng.sun@gmail.com}{jimeng.sun@gmail.com}}\\\\
${^1}$\uiuccs\\
${^2}$\iqvia\\
${^3}$\mgh\\
}
\title{SCRIB: Set-classifier with Class-specific Risk Bounds for Blackbox Models}

\begin{document}

\maketitle

\begin{abstract}
Despite deep learning (DL) success in classification problems, DL classifiers do not provide a sound mechanism to decide when to refrain from predicting. 
Recent works tried to control the overall prediction risk with \textit{classification with rejection options}. 
However, existing works overlook the different significance of different classes.
We introduce Set-classifier with Class-specific RIsk Bounds (\methodname) to tackle this problem, assigning multiple labels to each example.
Given the output of a black-box model on the validation set, \methodname constructs a set-classifier that controls the \textit{class-specific} prediction risks with a theoretical guarantee.
The key idea is to reject when the set classifier returns more than one label. 
We validated \methodname on several medical applications, including sleep staging on electroencephalogram (EEG) data, X-ray COVID image classification, and atrial fibrillation detection based on electrocardiogram (ECG) data. 
\methodname obtained desirable class-specific risks, which are 35\%-88\% closer to the target risks than baseline methods.
\end{abstract}

\section{Introduction}\label{sec:intro}
Deep Learning (DL) has demonstrated highly discriminative power on classification tasks and has been successfully applied in many application areas, including healthcare\cite{hannun2019cardiologist, Esteva2017-gs, Gulshan2016-bi, Biswal2018-lg}.

Impressive as DL is, we nevertheless hope to identify when the model might fail and take actions accordingly, which is especially important in healthcare applications. 
For example, suppose we are to design an automated system using pre-trained DL classifiers for sleep staging on EEG data~\cite{Biswal2018-lg}, detecting diseases based on  ECG data~\cite{Hong2019Mina:Signals}, or classifying X-ray images~\cite{Qiao2020Detection}.
For predictions to be reliable, the model should sometimes reject the examples and yield them to human experts to decide. And when the model does predict, we want the misclassification risks to be low and controllable.

This leads to classification with a reject option, where the rejection usually happens when the confidence score is low.
For example, when the base classifier's prediction is the true conditional probability, Maximum Class Probability (MCP) is the optimal confidence score as it minimizes the rejection rate for each risk level \cite{Chow1970OnTradeoff}.
The actual decision rule, given an overall risk target (not a class-specific one), \cite{Geifman2017SelectiveNetworks} picks a confidence threshold on the validation set.
Alternative confidence measures were also proposed by training separate models \cite{Jiang2018ToClassifier, Corbiere2019AddressingConfidence}.

However, existing works ignored that different classes have different significance. The confidence score is almost always class-agnostic, and the rejection is binary, which means there is no class-specific risk control.
As a result, a difficult class can have an extremely high rejection rate, where easy classes are predicted all the time. In many applications such as medicine, this class-agnostic rejection creates problems, as difficult classes are often the most important ones that need classification. 
For example, the N1 class in sleep staging is challenging to classify but of great interest to the applications. It will currently be disproportionally rejected due to the difficulty of achieving low overall risk. 

In this work, we aim to incorporate class-specific risk controls into classification with rejection.
We propose Set-classifier with Class-specific RIsk Bounds (\methodname), which can output multiple labels to each example based on the predicted conditional probabilities by a black-box classifier with theoretical guarantees.
Rejections happen naturally when the output set contains more than one label\footnote{We tackle  {\it multi-class classification} where one class is assigned to each example. This is different from {\it multi-label classification} where multiple labels can be assigned to the same example.}.
The multiple labels for each rejection also serve as an intuitive explanation of the underlying ambiguity, helping human experts understand the model behavior.

To construct the set classifier, \methodname searches for the optimal thresholds by minimizing a loss designed to control class-specific risks.
This set classifier can be optimal in some scenarios and naturally comes with a risk concentration bound. 
To the best of our knowledge, \methodname is the first class-specific risk control method for multi-class classification tasks with a theoretical guarantee. 
\methodname has the following desirable properties:
\begin{enumerate}
    \item {\bf Flexible}. It enforces class-specific risk controls by allowing different risk targets for different classes.
    \item {\bf General}. %It is \textit{post-hoc} and can work with any black-box classifier.
    It can work with any black-box classifier without model retraining.
    \item {\bf Concise}. When it rejects, it returns a set of possible labels, as the rejection explanation (Section \ref{sec:exp:user_study}) without unnecessary labels (Theorem \ref{thm:some_optimal}).
\end{enumerate}
Finally, we evaluated \methodname on multiple real-world medical datasets. 
\methodname obtained desirable class-specific risks (usually within 1\% of the target risks), which are 35\%-88\% closer to the target risks than baseline methods.

%========================================================================================================
\section{Related Works}\label{sec:related_works}

The most related line of works is a \textit{classification with rejection options}, which is intertwined with two other areas: \textit{calibration} and \textit{uncertainty quantification}. At a high level, classification with rejection options is about making classification decisions using a trained classifier, while calibration and uncertainty quantification enhance the classifier's prediction scores.

A natural method rejects if the prediction score (or uncertainty measure) is below (or above) a certain threshold.
In terms of the scores, one simple choice is the predicted class probabilities by the base classifier.
Many works directly use the predicted Maximum Class Probability (MCP) \footnote{In practice, usually MCP is replaced by the Maximum Softmax Response - the maximum Softmax output, as people tend to interpret Softmax output as probabilities.} \cite{Geifman2017SelectiveNetworks, Gimpel2017AExamples}, which is already optimal for overall risk control if the prediction is accurate \cite{Chow1970OnTradeoff}.
In this respect, \textit{calibration} research \cite{Platt1999ProbabilisticMethods, Guo2017OnNetworks, Wenger2020Non-ParametricClassification,Kull2019BeyondCalibration,Kumar2019VerifiedCalibration} is thus related as they aim to transform the classifier output to true probabilities. 
However, \textit{calibration} research is orthogonal to our problem - our work focuses on the decision (rejection) rules and does not require calibrated outputs.

Measures other than predicted probabilities have also been explored.
In classification, uncertainty is almost synonym to (the opposite of) confidence, and such research is related to \textit{uncertainty quantification}. 
Monte-Carlo Dropout (MCDropout) \cite{Gal2016DropoutLearning} is one of the most popular uncertainty quantification methods because it is relatively lightweight.
MCDropout was used in rejection literature \cite{Geifman2017SelectiveNetworks, Corbiere2019AddressingConfidence}.
However, most related methods (including MCDropout) \cite{Neal1996BayesianNetworks,Gal2016DropoutLearning, Blundell2015WeightNetworks,Wilson2016DeepLearning,Lakshminarayanan2017SimpleEnsembles, Moon2020Confidence-AwareNetworks, Corbiere2019AddressingConfidence} need to simultaneously train the base classifier and confidence/uncertainty estimator, which greatly limits the applicability and might even affect the performance, especially when the base classifier is a complicated deep learning model.
An exception is \cite{Jiang2018ToClassifier}, which trains a second classifier but is very expensive and only works for low dimensional data.
Works in {\it uncertainty quantification} are still complementary to our problem because uncertainty measures are inputs to the rejection rules, which will be demonstrated in our experiments.

Almost all score-based rejection works focus on finding better confidence measures and \cite{Geifman2017SelectiveNetworks, Fumera2000RejectThresholds} focus on decision rules (e.g., threshold finding). 
Apart from confidence-based rejection, a good number of works jointly learn the classifier and the rejector without an explicit confidence score at all—\cite{Fumera2002SupportOption_FIXED, Wegkamp2011SupportOption, Grandvalet2009SupportOption, Bartlett2008ClassificationLoss, Herbei2006ClassificationOption,Cortes2016BoostingAbstention,Cortes2016LearningRejection_FIXED, Geifman2019SelectiveNet:Option}, many of which focusing on binary classification and SVM.
Such methods also tend to have limited applicability and do not work for blackbox classifiers.

Most importantly, all works reviewed here focus on \textit{overall} risk.
Our work is the first to focus on finding the decision rules for class-specific risk controls given a blackbox classifier to the best of our knowledge.

A secondary issue of existing works is that the rejection is typically a binary decision. When rejection happens, we only know that the most likely class is selected or rejected. 
On the contrary, when our set-classifier rejects (i.e., when it contains more than one label), it informs the human inspector what competing predictions are causing the rejection given our risk targets (Section~\ref{sec:exp:user_study}).

%=========================================================================================================
\section{Problem Formulation}\label{sec:formulate}
\NotationTable
\subsection{Base Classifier and Learning Setup} \label{sec:formulate:learning_setup}
In this work we situate our task in the $K$-class classification problem, with data space $\mathcal{X}$, label space $\mathcal{Y} = \{1,\ldots,K\}$, and the joint distribution of $(X,Y)$ as $\mathbb{P}$ over $\mathcal{X}\times \mathcal{Y}$.
We will use $[K]$ to denote the set $\{1,2,\ldots,K\}$. 
We further denote the class-specific distributions $(X, Y=k)$ as $\mathbb{P}_k$, which is effectively a distribution over $\mathcal{X}$. Here
$\mathbb{P}_k\{\cdot\}$ can also be viewed as $\mathbb{P}\{\cdot|Y=k\}$.

Like in many tasks, we assume the data split into a training set $\trainingset$, validation set $\validationset$ and test set $\testset$. 
We will assume that data in $\validationset$ and $\testset$ follow the same distribution ($\mathbb{P}$) and are iid. And we can only use label information on $\trainingset$ and $\validationset$.
We are given a model $m$ (potentially a DNN) trained on $\trainingset$: $\mathcal{X} \mapsto \mathbb{R}^K$, where the $k$-th output, $m_k(x)$, captures the conditional probability $\mathbb{P}\{Y=k|X=x\}$. 
As a simple example,  $m(x)$ can be the Softmax output over the $K$ classes or confidence scores generated from uncertainty quantification or calibration methods. 
Note that here $\validationset$ is the validation set for this \textit{base classifier} $m$, and will be used to tune \methodname, as we will explain in Section \ref{sec:method}).
Finally, $\hat{\cdot}$ means evaluating the empirical value on $\validationset$. For example, $\hat{\mathbb{P}}\{Y=1\}$ means the frequency of class 1 in $\validationset$.

\subsection{Problem: Class-specific Risk Control}\label{sec:formulate:problem}
We will first introduce the concept of a set classifier:
\begin{definition}[\textbf{Set Classifier}] % \cite{Beran1987PrepivotingSets}%\cite{Sadinle2019LeastLevels}
    A set classifier is a mapping from data to a set of labels,  denoted as
    $\mathbf{H}:\mathcal{X} \mapsto 2^\mathcal{Y}$.
\end{definition}
A set-valued function has been used in classification tasks for different purposes (sometimes under different names) \cite{Wu2004ProbabilityCoupling, Vovk2005AlgorithmicWorld, DelCoz2009LearningClassifiers, Sadinle2019LeastLevels}.
Its link classification with rejection is straightforward:  Rejections happen naturally when the set classifier contains more than one label. 
A set classifier is a generalization of the typical classifier that only outputs the most-likely class.

For a multi-class classification problem, ideally, we want an oracle classifier such that $\mathbb{P}\{\mathbf{H}_{oracle}(X) = \{Y\}\} = 1$.
However, this is not possible in most cases. 
Our goal is to find a $\mathbf{H}$ that minimizes the \textit{ambiguity} while satisfying \textit{class-specific risk} constraints. 
There are many ways to define the ambiguity for a $\mathbf{H}$, and we will focus on two intuitive ones:

\begin{definition}[\textbf{Chance-Ambiguity and Size-Ambiguity}]\label{def:ambiguity}
    Chance-Ambiguity of a set classifier $\mathbf{H}$ is the probability of it having cardinality (size) greater than 1, namely $\mathbb{P}\{|\mathbf{H}(X)| > 1\}$.
    Size-Ambiguity is the expected size of $\mathbf{H}$, namely $\mathbb{E}[|\mathbf{H}(X)|]$
\end{definition}
These two ambiguity definitions of a set classifier $\mathbf{H}$ are usually correlated\footnote{Empirical results on the correlation are in the Appendix}.
Size-ambiguity is a measure used more often in the statistics literature \cite{Sadinle2019LeastLevels}, as it is easier to analyze.
However, it overlooks the qualitative difference between being certain ($|\mathbf{H}| = 1$) and uncertain ($|\mathbf{H}|>1$) - in reality, human experts usually need to get involved as long as the model is uncertain, regardless of the size of $\mathbf{H}$.
Chance-ambiguity is equivalent to the rejection rate widely used in rejection literature \cite{Geifman2017SelectiveNetworks, Jiang2018ToClassifier, Corbiere2019AddressingConfidence}.
We will use $A(\mathbf{H})$ to denote the general concept of ambiguity, either Chance-Ambiguity or Size-Ambiguity.

We define the class-specific risks as below.
\begin{definition}[\textbf{Class-specific Risk}]\label{def:classSpecRisk} 
The class-specific risk for a set classifier $\mathbf{H}$ for class $k$ is defined to be
\begin{equation}
r_k(\mathbf{H}) \defeq  \mathbb{P}_k\{k\not\in \mathbf{H}(X) \big| |\mathbf{H}| = 1\} \nonumber
\end{equation}
\end{definition}
Likewise, the overall risk for $\mathbf{H}$ is 
$$r(\mathbf{H}) \defeq  \mathbb{P}\{Y\not\in \mathbf{H}(X) \big| |\mathbf{H}| = 1\}$$ equivalent to the risk defined in existing rejection literature \cite{Geifman2017SelectiveNetworks}.

Intuitively, $\mathbb{P}_k\{k\not\in \mathbf{H}(X)\}$ means the probability of class $k$ not in the output of set classifier $\mathbf{H}(X)$ (or formally the {\it mis-coverage rate} of class $k$). And $\mathbb{P}_k\{k\not\in \mathbf{H}(X) \big| |\mathbf{H}| = 1\}$ is that probability of class $k$ for the output set with a single label (i.e., no ambiguity). 
Similarly, $\mathbb{P}\{Y\not\in \mathbf{H}(X) \big| |\mathbf{H}| = 1\}$ is such probability for all classes in general.

Putting everything together, our goal is to solve the following optimization problem:
\begin{problem}[\textbf{Class-specific Risk Control}]\label{eq:opt} 
\begin{align*}
    \min_{\mathbf{H}} \quad & A(\mathbf{H})\\
    \text{s.t. } \quad & \mathbb{P}_k\{k \not \in \mathbf{H}(X) \big| |\mathbf{H}(X)|=1\} \leq r^*_k, \forall k\in[K]
\end{align*}
\end{problem}
Here $A(\mathbf{H})$ is the ambiguity measure (Chance- or Size-Ambiguity, or a weighted average of both, chosen by the user depending on the task), and $r^*_1, \ldots, r^*_K \in[0,1]$ are the user-specified risk targets.
We seemingly ignored the case when $|\mathbf{H}| > 1$, but this constraint is implicit when we minimize $A(\mathbf{H})$. 
For example, the model's prediction is considered irrelevant when rejections happen in \cite{Geifman2017SelectiveNetworks}. 
For our method, the constraints can also be changed according to the task's specific goals (See Section \ref{sec:exp:exp_overall} for example).

\section{The \methodname Method}\label{sec:method}
\subsection{Method Overview}\label{sec:method:overview}

Given a trained base classifier $m$ and validation set $\validationset$,
we first parameterize $\mathbf{H}$ with $K$ thresholds, one for each class. 
 Given  $\mathbf{t}\defeq(t_1, \ldots,t_K)\in\mathbb{R}^K$, $\mathbf{H}(x)$ is defined as 
\begin{align}\label{eq:def_unionform}
    \mathbf{H}(x;\mathbf{t}) &\defeq \{k\in[K]: m_k(x)\geq t_k\}
\end{align}
In other words, we include $k$ in $\mathbf{H}(x)$ if the model thinks the likelihood of $x$ belonging to class $k$ denoted as $m_k(x)$  is above a threshold $t_k$.

Figure \ref{fig:explain_set} illustrated an example when $K=3$.
Here $m_k(x)$ is a proxy for how confident the model thinks $x$ is from class $k$.
We will drop $\mathbf{t}$ in the notation for simplicity.
\FigExplainSetClassifier

Next, we transform the optimization problem in Section \ref{sec:formulate:problem} into an unconstrained optimization problem that minimizes the following loss $\hat{L}: \mathbb{R}^K \mapsto \mathbb{R}$:
\begin{align}\label{eq:loss}
    \hat{L}(\mathbf{t}) \defeq \underbrace{\hat{A}(\mathbf{H})}_{\text{ambiguity}} + \sum_{k=1}^K \underbrace{\lambda_k( \hat{r}_k(\mathbf{H}) - r_k^*)_{+}^2}_{\text{class specific risk control penalty}}
\end{align}
where $v_{+}\defeq \max\{v,0\}$ and $\mathbf{H}$ is parameterized by $\mathbf{t}$ as defined above.
$\hat{A}$ and $\hat{r}_k$ are the ambiguities and risks evaluated on the validation set $\validationset$.
Obviously, one can use the same penalty coefficient for all classes - namely $\lambda_k \equiv \lambda$ for a fixed $\lambda$ unless there is a strong prior certain class is more important than others.

Although the loss definition seems simple, the interaction between different classes makes it hard to simultaneously optimize all parameters.
To tackle the actual optimization, we choose the thresholds $\mathbf{t}$ from the model's outputs on the validation set $\validationset$.
To this end, we tried Bayesian Optimization and a variant of coordinate descent algorithm that optimizes only one $t_k$ at a time.
Empirically we find that the coordinate descent algorithm works much better.
Full details of the algorithm are provided in Algorithm \ref{alg:main}.
\AlgoMain
In practice, we repeat Algorithm \ref{alg:main} ten times and take the lowest loss found.

\textbf{Complexity}
The naive search in each direction requires $O(N)$ loss evaluations, which each takes $O(KN)$.  It leads to $O(KN^2)$ operations.
We used a dynamic programming trick in QuickSearch, lowering it to $O(KN)$ operations instead. 
Total complexity is thus $O(TK^2N)$ where $T$ denotes the number of outer-iterations in Algorithm \ref{alg:main}. 

\textbf{Implementation trick} To further speed up, one can also search for $\mathbf{t}$ only on a subset of $\validationset$.
In our experiments, the optimization usually only takes up to a few seconds, possibly because $K$ is relatively small. 
When $K$ is big, it only makes sense to manually choose the target risk for a relatively small number of classes - either because they are very important or very difficult or easy.
The rest of the classes should share the same threshold as a corollary of Theorem \ref{thm:some_optimal}.
This will effectively reduce the problem of searching a small number of thresholds as well.
Due to the space constraint, we have the pseudo-code for searching the minimum in each coordinate and comparison with several plug-in optimization methods (time and value) in the Appendix.

\subsection{Parameterization Optimality}\label{sec:method:set_theory}
By parameterizing $\mathbf{H}$ using $\mathbf{t}$ as in Eq. (\ref{eq:def_unionform}), we are answering the question ``Might $x$ belong to class $k$?'' for each class separately, as illustrated in Figure \ref{fig:explain_set}.
It seems this particular parameterization ``ignored'' the potential interaction between classes. 
However, as we will prove next, $\mathbf{H}$ is already optimal in minimizing the mis-coverage rate. 
Here we define the \textit{mis-coverage rate} for $\mathbf{H}$ for class $k$ as:
\begin{align}\label{eq:classSpec}
    \alpha_k(\mathbf{H}) \defeq \mathbb{P}_k\{k\not\in \mathbf{H}(X)\}
\end{align}
It refers to the probability that the correct class $k$ is not in the output of set classifier $\mathbf{H}(X)$. 

\begin{theorem}\label{thm:some_optimal} (Adapted from \cite{Sadinle2019LeastLevels})
For any $\mathbf{t}$, define $\mathbf{H}^*$ as the set classifier parameterized by $\mathbf{H}^*(x) \defeq \{k: \mathbb{P}\{Y=k|X=x\} > t_k \}$. 
$\mathbf{H}^*$ has the minimum Size-Ambiguity among all set classifiers with equal or lower mis-coverage rates.
That is, $\forall \mathbf{H}'$ 
$$
\big(\forall k, \alpha_k(\mathbf{H}') \leq  \alpha_k(\mathbf{H}^*)\big) \Leftrightarrow \mathbb{E}[|\mathbf{H}^*|] \leq \mathbb{E}[|\mathbf{H}'|]
$$
\end{theorem}
A proof using the Neyman-Pearson lemma \cite{Neyman1933IX.Hypotheses} is included in Appendix.

Usually (and in all our experiments), the base classifier gives us some prediction scores (e.g., Softmax output). 
Un-calibrated prediction scores tend to deviate from true probabilities \cite{Guo2017OnNetworks}, but we do not need $m_k(x)$ to be close to $\mathbb{P}\{Y=k|X=x\}$. 
Instead, we only need order consistency - that is, $\forall x,x',k$, $m_k(x)>m_k(x') \Leftrightarrow \mathbb{P}\{Y=k|X=x\} > \mathbb{P}\{Y=k|X=x'\}$. 
%As most calibration method is order-preserving, \methodname can use uncalibrated output directly as in Eq. (\ref{eq:def_unionform}). 
If the base classifier captures the ordering of $\mathbb{P}\{Y=k|X=x\}$, then with Theorem \ref{thm:some_optimal}, our parameterization in Eq. (\ref{eq:def_unionform}) will give us an optimal $\mathbf{H}$ for minimizing Size-Ambiguity.

When the objective function contains Chance-Ambiguity, the form of $\mathbf{H}$ will depend on the distribution of the predictions (assuming they are true conditional probabilities). 
However, our proposed parameterization is still desirable because, empirically, Chance- and Size-Ambiguity are correlated, and this simple parameterization is also intuitive and less prone to over-fitting.

\noindent{\bf Secondary output:}
Another benefit of this parameterization is that for each output $\mathbf{H}$, we have the estimated mis-coverage rates $\alpha_1(\mathbf{H}),\ldots, \alpha_K(\mathbf{H})$ immediately\footnote{This is given by the quantiles of the thresholds $\mathbf{t}$.
}. 
Intuitively, the mis-coverage rate means $\mathbf{H}$ can miss class $k$ with only probability $\alpha_k(\mathbf{H})$. 
As output, $\alpha_k(\mathbf{H})$ can be beneficial for human experts in classifying the rejected samples. 

\subsection{Risk Bounds}\label{sec:method:concentrate_theory}
As mentioned in Section \ref{sec:method:overview}, the thresholds $\mathbf{t}$ are chosen based on the empirical loss on the validation set $\validationset$, which is equivalent to enforcing the risk constraints in Problem \ref{eq:opt} on $\validationset$.
By selecting $\mathbf{t}$ on $\validationset$, we borrow ideas from the Split Conformal method \cite{Vovk2005AlgorithmicWorld, Lei2018Distribution-FreeRegression, Papadopoulos2002InductiveRegression_FIXED}: 
Because the model is not trained on the validation data nor the (unseen) test data, if data in $\validationset$ and $\testset$ follow the same distribution, then the scores' distribution on $\validationset$ for each class $k$ can represent that at test time.

Denoting the validation set as $\{(X_1,Y_1),\ldots,(X_N, Y_N)\}$.
For a \textit{fixed} set classifier $\mathbf{H}$ parameterized by the base classifier and thresholds $\mathbf{t}$,  if we evaluate its risks ($\hat{r}_k(\mathbf{H})$) on this validation set that the base classifier was not trained on, then for a new data $X_{N+1}$ at test time, $\mathbf{H}(X_{N+1})$ will still in expectation have the same risk as $\hat{r}_k(\mathbf{H})$:
\begin{theorem}\label{thm:modified_risk_still_good_in_expectation} 
For any \textit{fixed} set classifier $\mathbf{H}$ parameterized by $\mathbf{t}$, and new data $\{(X_i, Y_i)\}_{i=1}^{N+1}$ from $\mathbb{P}$, denote $k = Y_{N+1}$ as the true class of $X_{N+1}$, we have
$$
 \mathbb{P}_k\{k\not\in \mathbf{H}(X_{N+1}) \big||\mathbf{H}(X_{N+1})|=1\} =\mathbb{E}[\hat{r}_k(\mathbf{H})] = r_{k}(\mathbf{H})
$$
where $\hat{r}_k(\mathbf{H})$ is the risk on the first $N$ data points and $r_k(\mathbf{H})$ is the true risk defined in Definition \ref{def:classSpecRisk}.
Moreover, with \cite{Hoeffding1963ProbabilityVariables} we have $\forall \epsilon > 0$: 
\small $$
\mathbb{P}\{\hat{r}_k(\mathbf{H}) \geq r_k(\mathbf{H}) + \epsilon \}  \leq e^{-D(r_k(\mathbf{H})+\epsilon ||  r_k(\mathbf{H})) n_k} 
$$
%https://en.wikipedia.org/wiki/Chernoff_bound , Additive form (absolute error)
$$\text{where }
D(p||q) = p\ln{\frac{p}{q}} + (1-p)\ln{\frac{1-p}{1-q}}
$$ 
is the Kullback–Leibler divergence between Bernoulli random variables parameterized by $p$ and $q$, 
and $n_k \defeq \sum_{i=1}^N \mathbf{1}\{Y_i=k\}\mathbf{1}\{|\mathbf{H}(X_i)|=1\}$ denotes the number of data points from class $k$ that receives a certain prediction by $\mathbf{H}$.
\end{theorem}
Proof for Theorem \ref{thm:modified_risk_still_good_in_expectation} is in Appendix.

\section{Empirical Results}\label{sec:exp}
We present a few closely relevant baselines to our task in Section \ref{sec:exp:baselines}.
We will then compare these methods (when applicable) to \methodname on a series of risk control tasks on synthetic and real-world datasets. The real-world datasets are with diverse characteristics but all from the medical domain, because we believe {\it classification with rejection} can have an important practical impact on that domain.

\subsection{Baselines}\label{sec:exp:baselines}
We compare \methodname with the following baselines. 
\begin{itemize}
\item \textbf{Selective Guaranteed Risk (SGR)} \cite{Geifman2017SelectiveNetworks}
is a post-hoc method that can achieve an overall risk guarantee. 
The proposed version uses (predicted) Maximum Class Probability of the base classifier as the confidence score for rejection.

\item \textbf{SGR + Dropout} \cite{Geifman2017SelectiveNetworks} is a variant of SGR using the (negative) variance of Monte-Carlo Dropout \cite{Gal2016DropoutLearning} predictions as the confidence score. 

\item \textbf{LABEL} \cite{Sadinle2019LeastLevels} is a set-classifier that can control the class-specific \textit{mis-coverage rate} $\alpha_k(\mathbf{H})$, the unconditional version of $r_k(\mathbf{H})$. 
It uses an analytical solution specific to $\alpha_k(\mathbf{H})$.

\item \textbf{\methodname-} The same as \methodname but we use the same threshold for all classes $t_k\equiv t$ for the same $t$. 
We include this to check the necessity of using multiple thresholds. 

\end{itemize}
Compared with SGR, \methodname can provide class-specific risk controls along with additional information (a confidence set) to human decision-makers when rejections happen.
Compared with LABEL, \methodname can control both the unconditional coverage level (as a degenerate use case, see Appendix) and the conditional risk when $|\mathbf{H}| = 1$. 
In addition, we want to emphasize that \methodname can be applied to solve a lot of \textit{more general} problems, with the specific optimization in Eq. (\ref{eq:opt}) being just an instance.
As an example, we will explain how \methodname can be modified mildly to control overall risk in Experiment \ref{sec:exp:exp_overall}.

\subsection{Data and Model Output}\label{sec:exp:data}
\textbf{Synthetic} data is created by first generating conditional probabilities and then sampling the labels from these probabilities. 
Synthetic data is helpful because we can evaluate the risk control methods independent of the underlying base classifier. 
The synthetic data has 5 classes, with an easy class and a hard one.
The exact details are in the Appendix. 

\textbf{ISRUC} \cite{Khalighi2016ISRUC-Sleep:Researchers} (Sub-group 1) is a publicly available PSG dataset for sleep staging task.
The sub-group 1 data contains PSG recordings of 100 subjects (89,283 examples\footnote{subject \# 8 is excluded due to missing channels}) sampled at 200Hz, and we use the 6 EEG channels (F3, F4, C3, C4, O1, and O2).
75\% of the data are used for training the base classifier, and the rest is split evenly into validation and test sets.
Sleep stage labels are assigned every epoch (30 seconds).
Class 0-4 are W/N1/N2/N3/REM, respectively.

\textbf{Sleep-EDF} \cite{Kemp2000AnalysisEEG, Goldberger2000PhysioBankSignals.} is another public dataset widely used to evaluate sleep staging models.
The version we used as of the end of 2020 contains 153 whole-night Polysomnographic (PSG) recordings of 2 channels (Fpz-Cz and Pz-Oz) at 100Hz. 
We use 122 recordings (331,184 samples) for training the base classifier and evenly split the rest into validation and test sets. 
The sleep stage labels are assigned continuously using start and end times.
 Class 0-4 are W/N1/N2/N3/REM, respectively.
%The number of W (Wake) classes are significantly more than ISRUC because the recording last over the day time as well.
%https://arxiv.org/pdf/1910.11162.pdf
%https://physionet.org/content/sleep-edfx/1.0.0/

\textbf{ECG (PhysioNet2017)} \cite{Clifford2017AF2017, Goldberger2000PhysioBankSignals.} is a publicly available ECG dataset with 8,528 de-identified ECG recordings sampled at 300Hz, containing Normal (N), Atrial Fibrillation (AF), Other rhythms (O), and Noisy recordings. 
75\% of the recordings were used for training, and the rest were evenly split into validation and test sets. 
Class 0-3 are N/O/AF/Noisy, respectively.

\textbf{X-ray} dataset is constructed from two publicly available sources, COVID Chest X-ray\footnote{https://github.com/ieee8023/covid-chestxray-dataset} and Kaggle Chest X-ray\footnote{https://www.kaggle.com/paultimothymooney/chest-xray-pneumonia}, including 5,508 chest X-ray images from 2,874 patients.
Class 0-3 are  COVID-19, non-COVID-19 viral pneumonia, bacterial pneumonia, and normal.

Excluding samples for model training, each class's sample counts for each validation dataset are presented in Table \ref{table:exp:data:cnts}.
\TabDataClassCounts

\textbf{Base Deep Learning Models}
For ISRUC and Sleep-EDF, we used a ResNet-based \cite{He2016DeepRecognition} with 3 Residual Blocks, each with 2 convolution layers.
It first performs a Short-time Fourier transform (STFT) on the data, passes the output through a convolutional layer, the Residual Blocks, and then 2 fully connected layers. 
For ECG, we employed \cite{Hong2019Mina:Signals} and changed the last layer for a 4-classification problem. 
For experiments with SGR+Dropout baseline, we add a dropout layer before weights except for the input per \cite{Gal2016DropoutLearning}. 
For X-ray, we directly take the DL model predictions \cite{Qiao2020Detection} and run experiments in a purely post-hoc manner. %https://www.ncbi.nlm.nih.gov/pmc/articles/PMC7665533/pdf/ocaa280.pdf
More training details are in the Appendix.

\subsection{Experiments}
In the experiments, we aim to answer the following questions:
\begin{itemize}
    \item Can \methodname control class-specific risks well empirically? (Section \ref{sec:exp:exp_classspec})
    
    \item Does \methodname also perform well for overall risk control? (Section \ref{sec:exp:exp_overall}) 
    This experiment also serves as a test for our optimization method.
    
\end{itemize}

\subsubsection{Experiment: Class-Specific Risks} \label{sec:exp:exp_classspec}
In this experiment, we want to see if \methodname can find a $\mathbf{H}$ with a risk profile similar to a set of pre-specified values. 
In many healthcare-related tasks, some classes are (much) harder to classify than others.
For example, in sleep staging datasets, N1 is usually the hardest-to-predict class, whereas W (wake) is usually easy, which means the risks are very high/low risk for N1/W. 
Figure \ref{fig:exp2:arb} illustrates this observation and shows how potentially we could set the risk targets to alleviate this issue with \methodname.
\FigClassSpecArbitrary

\textbf{Setup}:
To quantitatively compare different methods, we will set the target risks ($r_k^*$) for \methodname to 15\% for all classes for ECG and 10\% for other datasets.
Same numbers are used as overall risk targets ($r^*$) for SGR and mis-coverage targets for LABEL.
The target is higher for ECG because the performance of the classifier is worse (SGR already rejects 90+\% samples at $r^*=15\%$).
$\lambda_k$ is set to $10^4$ for all classes and datasets, and we use chance-ambiguity for $A(\mathbf{H})$.
We choose large $\lambda_k$s to satisfy the risk constraint before optimizing ambiguities
 (see Eq.~\ref{eq:loss}). 
In fact, $10^4$ is not that large, as 1\% excess risk translates to $10^4 (1\%)^2=1.0$ (the second term in Eq.~\ref{eq:loss}), while the ambiguity term (the first term) is a value in $[0,1]$.

We repeat the experiment 20 times, each time randomly re-splitting unseen data evenly into validation and test sets.
For ISRUC/SleepEDF/ECG, we include the results by re-splitting recordings/subjects in the Appendix. 

\FigTabExpII
\textbf{Evaluation Metric}:
We will measure the excess class-specific risk
$$
(\Delta r_k)_+ \defeq\max\{0, r_k(\mathbf{H}_{method}) - r_k^*\}
$$
on the test set, where $method$ can be SGR\footnote{For binary rejection like SGR, $\mathbf{H}_{SGR}(x)$ is naturally defined to be $[K]$ when rejections happen.}, LABEL, \methodname- and \methodname.

\textbf{Results} are presented in Table \ref{table:exp_classSpec} and Figure \ref{fig:exp2}.  
The runtime of \methodname is detailed in the Appendix, which is generally a few seconds.
\methodname almost always controls the class-specific risks close to the target. 
Except for the X-ray dataset, the difference between \methodname and the best baseline is always significant. 
This can also be seen from the violin-plots as well.
For the X-ray dataset, the risks are much more volatile as each class size is small, especially after rejection.
Comparison between \methodname and \methodname- suggests that using the same threshold for all classes is not enough even with the custom loss function. 

\subsubsection{Clinical User Study of Set Predictions} \label{sec:exp:user_study}
To evaluate the practical value and interpretability of a set classifier, we picked 50 samples from the ISRUC dataset\footnote{For each class, we pick the most certain instance according to the base classifier, 3 instances at the 100\%/90\%/80\% percentile for entropy, and 6 purely random instances.} and asked a neurologist with a specialization in sleep medicine to score the predicted sets from 1 to 5 (with 5 being the best).
The sets get lower scores if they miss a likely class or unnecessarily ambiguous (e.g., contain all labels all the time). 
\FigUserViolin
We compare the scores with a baseline that uses the same $t$ for all classes like \methodname- and SGR, where $t$ is chosen to have the same number of certain predictions as \methodname.
On average, \methodname's score is significantly higher with p-value 0.01 ($3.86\pm 0.86$ vs $3.42\pm 0.91$).

\subsubsection{Experiment: Overall Risk} \label{sec:exp:exp_overall}
This experiment focuses on comparing the overall risk control between \methodname and the baseline method SGR.
The first goal is to explain how to slightly change the loss function of \methodname for a different task, such as the overall risk control SGR was designed for.
Moreover, because we know the analytic solution when the predicted probabilities are accurate, this experiment also serves as a sanity check to see whether the searched local optima are good (close to global optima).

\textbf{Setup}:
We will use \methodname to solve the overall risk control SGR was designed for, by changing the loss function to account for chance-ambiguity and the overall risk:
\begin{align}\label{eq:loss:overall}
    \hat{L}_{overall}(\mathbf{t}) &\defeq \underbrace{\hat{\mathbb{P}}\{|\mathbf{H}(X)|>1\}}_{\text{Chance-ambiguity}} + \underbrace{\lambda( \hat{r}(\mathbf{H}) - r^*)_{+}^2}_{\text{Overall risk penalty}}
\end{align}
Note that setting all thresholds to the same gives the best trade-off when the base classifier is accurate, but we do not impose this prior knowledge.
Therefore, an inferior search could find bad local optima/trade-offs for \methodname because it picks $K$ different thresholds.
We repeat the experiment 20 times, each time randomly re-splitting unseen data evenly into validation and test sets.
For ISRUC/SleepEDF/ECG, data for the same patient are always in the same set.
$\lambda$ is set to $10^4$ like before.

\textbf{Evaluation Metric}:
We will plot accuracy ($1 -$ risk) as a function of coverage / chance-ambiguity and compute the area under the curve (AUC) for SGR and \methodname. 
This is the common evaluation metric in classification with rejection literature.
When the model output is the true conditional probability, using the same threshold $t$ for all classes is theoretically optimal. 
As a result, we expect the SGR curve to be above \methodname for the Synthetic data (i.e., lower ambiguity with the same risk), but not too much.
\FigTabExpI

\textbf{Results} are presented in Table \ref{table:exp_overall} and Figure \ref{fig:exp1}.
In general, \methodname is on par with or better than SGR in our benchmark datasets.
Although SGR is the theoretical optimal on the Synthetic data, the performance difference between \methodname and SGR is small. 
This is also the case for Xray, but for the rest of the data, we see that \methodname has the best trade-off.
This is a known phenomenon \cite{Fumera2000RejectThresholds} and can happen if the base classifier has biases for a different class. But the focus of this experiment is that our search algorithm finds good local optima. 
SGR+Dropout is comparable with SGR. 
On ECG, the confidence given by MCDropout is negatively correlated with accuracy, which prevents SGR from controlling the overall risk, so we omit those results\footnote{There is no curve because it can never find a threshold such that data \textit{above} that threshold have a low risk. 
Similar phenomena have been noted before \cite{Jiang2018ToClassifier}}.

\section{Conclusion}
In this paper, we present \methodname, the first method for classification with rejection with class-specific risk controls.
\methodname provides a simple and effective way to construct set-classifiers for this task by choosing multiple thresholds for the base classifier's output.
We demonstrated how overall risk control leads to the issue of unbalanced risks for different classes.
Then, we showed that \methodname can control the class-specific risks close to the targets on several medical datasets.
\methodname has potential applications to other fields where class-specific risks matter as well.

%===================================================================Appendix
\appendix
\onecolumn
%\paragraph{Appendix}
%In this supporting material, we first present the proofs for the two theorems in the main text in Section \ref{appendix:sec:proof}.
%We will then go over the specific search algorithm we used in each iteration of the coordinate descent in Section \ref{appendix:sec:impl}.
%Finally, we will present more details on the experiments that we cannot fit in the main text in Section \ref{appendix:sec:exp}, which also includes additional details comparing the correlation between Chance- and Size-Ambiguities, comparison with LABEL, and comparison of different optimization methods for our loss function. 

\section{Appendix: Proofs}\label{appendix:sec:proof}

\subsection{Proof for Theorem 4.1}

Denote $\pi_k = \mathbb{P}\{Y=k\}$. 
For any set classifier $\mathbf{H}$, we have 
\begin{align}
    \mathbb{E}[|\mathbf{H}|] &= \sum_{k=1}^K\mathbb{E}[\mathbf{1}\{k \in \mathbf{H}(X)\}] = \sum_{k=1}^K\mathbb{P}\{k \in \mathbf{H}(X)\}\\
    %&= \sum_{k=1}^K \sum_{y=1}^K \mathbb{P}_y\{k \in \mathbf{H}(X)\}\pi_y\\
    &= \sum_{k=1}^K \mathbb{P}_k\{k\in \mathbf{H}(X)\} \pi_k + \sum_{k=1}^K \sum_{y\neq k} \mathbb{P}_y\{k\in \mathbf{H}(X)\} \pi_y \\
    &= \underbrace{\sum_{k=1}^K (1-\alpha_k(\mathbf{H})) \pi_k}_{\text{``Necessary'' inclusion}: N(\mathbf{H})} + %\underbrace{\sum_{k=1}^K \mathbb{P}\{k\in \mathbf{H}(X), Y\neq k\}}_{\text{``Unnecessary''}: U(\mathbf{H})}
    \underbrace{\sum_{k=1}^K \sum_{y\neq k} \mathbb{P}_y\{k\in \mathbf{H}(X)\}\pi_y}_{\text{``Unnecessary''}: U(\mathbf{H})}
\end{align}
Suppose we have $\mathbf{H}$ and $\mathbf{H}'$ as described in Theorem 3.1, and let's denote the ``Necessary'' and ``Unnecessary'' inclusions of $\mathbf{H}$ as $N(\mathbf{H})$ and $U(\mathbf{H})$. 
We already have $N(\mathbf{H})\leq N(\mathbf{H}')$ by the definition of $\mathbf{H}'$, so we just need to show that $U(\mathbf{H})\leq U(\mathbf{H}')$. 

\bigbreak
We introduce the Neyman-Pearson Lemma before further discussion:
\begin{definition} (Likelihood-ratio test)
For two hypotheses $H_0:\theta=\theta_0$ and $H_1:\theta\neq \theta_0$, the likelihood-ratio test at significance level  $\alpha$ rejects $H_0$ if and only if the likelihood ratio $\frac{\mathcal{L}(H_0|x)}{\mathcal{L}(H_1|x)} < t$ for some threshold $t$.
Here $\alpha$ is the type I error rate $\mathbb{P}\{\text{reject} |H_0\}$, and the type II error rate is defined to be $\beta\defeq\mathbb{P}\{\text{fails to reject} | H_0\}$.
\end{definition}

\begin{lemma} (\cite{Neyman1933IX.Hypotheses})
The likelihood test is the most powerful test that rejects $H_0$ in favor of $H_1$ at significance level (equal to or less than) $\alpha$. 
In other words, it has the minimum type II error rate $\beta$ compared with any other test with the same $\alpha$.
\end{lemma}
Consider the null hypothesis $H_{0,k}: Y=k$ vs the alternative $H_{1,k}:Y\neq k$.
Since 
$$
\frac{\mathcal{L}(H_{0,k}|x)}{\mathcal{L}(H_{1,k}|x)} = \frac{\mathbb{P}\{X=x|Y=k\}}{\mathbb{P}\{X=x|Y\neq k\}} = \frac{1-\pi_k}{\pi_k}\frac{\mathbb{P}\{Y=k|X=x\}\mathbb{P}\{X=x\}}{\mathbb{P}\{Y\neq k|X=x\}\mathbb{P}\{X=x\}} = g(\mathbb{P}\{Y=k|X=x\})
$$
with $g(x)=\frac{1-\pi_k}{\pi_k}\frac{x}{1-x}$ being a monotonic function, choosing a threshold for likelihood-ratio is the same as choosing a threshold for $\mathbb{P}\{Y=k|X=x\}$. 
Therefore, the test ``rejects $H_{0,k}$ iff $k\not \in \mathbf{H}(x)$'' is equivalent to the likelihood-ratio test.

In this case we have $\beta_k =\mathbb{P}\{k\in \mathbf{H}(X)|Y\neq k\}$ and $\alpha_k = \mathbb{P}_k\{k\not\in\mathbf{H}(X)\}$.
By our assumption, $\forall k, \alpha_k(\mathbf{H}')\leq \alpha_k(\mathbf{H})$, so Neyman-Pearson Lemma means that 
\begin{align*}
    \forall k\in[K], \beta_k(\mathbf{H}') &\geq \beta_k(\mathbf{H})\\
    \implies \sum_{k=1}^K (1-\pi_k)\beta_k(\mathbf{H}') &\geq \sum_{k=1}^K (1-\pi_k)\beta_k(\mathbf{H})\\
    \implies \sum_{k=1}^K\sum_{y\neq k}  \mathbb{P}\{k\in \mathbf{H}'(X), Y= y\} &\geq \sum_{k=1}^K\sum_{y\neq k}  \mathbb{P}\{k\in \mathbf{H}(X), Y= y\}\\
    \implies \sum_{k=1}^K\sum_{y\neq k}  \mathbb{P}_y\{k\in \mathbf{H}'(X)\}\pi_y &\geq \sum_{k=1}^K\sum_{y\neq k}  \mathbb{P}_y\{k\in \mathbf{H}(X)\}\pi_y\\
    \implies U(\mathbf{H}') &\geq U(\mathbf{H})
\end{align*}

\subsection{Proof for Theorem 4.2}
First, we prove the expectation. Denote event $A$ as ``$k\not\in\mathbf{H}(X) \land |\mathbf{H}(X)|=1$'', and $B$ as ``$|\mathbf{H}(X)|=1$'', and $N_E$ as the number of times the event $E:=A$ or $E := B$ happens on the validation set, we have
\begin{align*}
    &E[\frac{N_A}{N_B} | N_B\geq 1]\\
    = &\sum_{m=1}^N E[\frac{N_A}{m} | N_B=m]\mathbb{P}\{N_B=m | N_B>=1\} \text{ (Partition Theorem / Law of Total Probability)}\\
    =&\sum_{m=1}^N \mathbb{P}\{A|B\}\mathbb{P}\{N_B=m | N_B>=1\} \text{ (Definition of $\mathbb{P}\{A|B\}$ )}\\
    = &\mathbb{P}\{A|B\} = R_k(\mathbf{H})
\end{align*}
Since all $(X_i, Y_i)$ are iid, we have $\mathbf{P}_k\{k\not\in\mathbf{H}(X_{N+1}) | |\mathbf{H}(X_{N+1})| = 1\} = R_k(\mathbf{H})$ as well. 

It should be clear that $A|B$ follows a Bernoulli distribution at this point (this is the same idea as negative sampling). 
The concentration bound, namely the Hoeffding inequality, follows directly. 

\textbf{Remarks}
Note that if $\mathbf{H}$ was optimized on $\validationset$ to minimize the risks, then Theorem 4.2 does not directly apply.
The analogy is that the test error should be higher than the training error. 
Our method does \textit{not} minimize the risks, but we do have a search step that will prevent us from using the validation set to compute the concentration bound. 
A practical solution to this would be first tuning $\mathbf{H}$ on the validation set and then use the test set (or a second validation set) for the concentration bound. 
In reality, when we have enough data ($N\gg K$), $\hat{r}(\mathbf{H})$ on the first validation set is usually already very close to the true risk.

\section{Appendix: Implementation Details of Coordinate Descent Search} \label{appendix:sec:impl}
\AlgoSearch
Denote the size of the validation set as $N$. 
Given the specific loss function we have, fixing the thresholds for other classes, we can evaluate all potential thresholds for one class in $O(KN)$ time - the same complexity as just one loss evaluation. 
WLOG, suppose we are fixing $\{t_k\}_{k\neq d}$ and optimizing $t_d$. 
For each loss evaluation, we need ambiguity $\hat{A}(\mathbf{H})$ and $\hat{R}_k(\mathbf{H})$ for each $k\in[K]$.
Denote $\mathbf{M}_{i,k}$ as the $i$-th smallest value in $\{m_k(x)\}_{x\in\validationset}$, and $\mathbf{I}_{i,k}$ as the original index (i.e. $m_k(x_{\mathbf{I}_{i,k}}) = \mathbf{M}_{i,k}$).
We book-keep the following values while we iterate over $j\in[N]$ and set $t'_d$ to $\mathbf{M}_{j,d}$:
\begin{itemize}
    \item for each $i\in[N]$, denote the size of $\mathbf{H}(x_i)$ if $t_d$ is set to $\mathbf{M}_{j, d}$ as:
    $$
    cnt_{i,j}\defeq \underbrace{|\{k\in[K]\setminus\{d\}: m_k(x_i) > t_k\}|}_{\text{labels in $\mathbf{H}(x_i)$ that is not $d$}} 
    + \underbrace{\mathbf{1}\{m_d(x_i) > \mathbf{M}_{j,d}\}}_{\text{whether $d$ is in $\mathbf{H}(x_i)$ }}
    $$
    
    \item for each $k\in[K]$, denote the number of \textit{certain} predictions that belong to class $k$ as:
    $$sure_{k,j}\defeq |\{i: y_i=k, cnt_{i,j} = 1\}|$$
    
    \item for each $k\in[K]$, denote the number of \textit{certain but wrong} predictions that belong to class $k$:
    $$err_{k,j}\defeq |\{i: m_k(x_i) \leq t_k, y_i=k, cnt_{i,j} = 1\}|$$
\end{itemize}
Chance-ambiguity\footnote{Size-ambiguity (denoted as $A_{S}$) is computed simply as $A_{S,j} \gets A_{S, j-1} - \frac{1}{N}$.} (denoted as $A_{C}$) is then computed as $A_{C,j} \gets 1 - \frac{\sum_{k\in[K]}sure_{k,j}}{N}$.
The risk penalty is computed as $R_{k,j} \gets \frac{err_{k,j}}{sure_{k,j}}$. 
Note we would not want to choose $t_d = \mathbf{M}_{N, d}$, as that means we never predict class $d$. 
Putting everything together, we have the final algorithm in Algorithm \ref{alg:search_each_iter} (QuickSearch).

\textbf{Complexity} The routine QuickSort has $O(N)$ iterations, each taking $O(K)$ (for the compution of loss). 
The initialization also takes $O(KN)$ time, so the total runtime is $O(KN)$. 
Note that the sorting of model output $\mathbf{M}$ and indices $\mathbf{I}$ only happens once and can be re-used for each QuickSearch.

\section{Appendix: Additional Experiment Results}\label{appendix:sec:exp}
Codes used in this paper are available at
\href{https://anonymous.4open.science/r/6789ca6b-3c4f-4c7f-a859-f7cdf2c11198/}{https://anonymous.4open.science/r/6789ca6b-3c4f-4c7f-a859-f7cdf2c11198/}

\subsection{Training of Deep Learning Models}

All learning is implemented using PyTorch \cite{Paszke2019PyTorch:Library}.
For clarity we will use the PyTorch layer name in the descriptions (e.g. \texttt{Conv2d} for 2D convolutional layer).

\textbf{SleepEDF/ISRUC}: 
For the model, we employ a ResNet-based architecture \cite{He2016DeepRecognition}.
\begin{enumerate}
    \item The model first perform a short time fourier transform on the input data using \texttt{torch.stft} with size \texttt{n\_fft} set to 256, \texttt{hop\_length} set to 64 and no padding on both sides (\texttt{center=False}). 
    
    \item The result is then passed through \texttt{Conv2d-BatchNorm2d-ELU} sequentially. 
    Conv2d has $3c$ filters of size 3, with \texttt{stride=1} and \texttt{padding=1}, where $c$ is the number of channels (6 for ISRUC and 2 for SleeepEDF).
    
    \item The result is passed through 3 ResBlock consecutively, with $4c$, $8c$, $16c$ filters, respectively. 
    Each ResBlock uses \texttt{Conv2d-BatchNorm2d-ELU-Conv2d-BatchNorm2d} for the residual learning.
    Kernel size is always 3, and stride is set to 2. 
    After merging residual with a downsampled input, the result is passed through a \texttt{Dropout} layer with a probability of 0.5
    The last 2 ResBlocks perform a stride 2 \texttt{MaxPool2d} before \texttt{Dropout}. 
    
    \item Finally, the result is passed through a fully connected layer with $16c$ nodes and another with K=5 read-out nodes. 
    
\end{enumerate}
Following the ISRUC's own label assignment convention, we split the original recordings into 30-second epochs for the data processing.
For ISRUC, we use the 6 EEG channels (F3, F4, C3, C4, O1, and O2). 
ISRUC has been downsampled to 100Hz so we can share the exact model between the two datasets.
For SleepEDF, we use the 2 EEG channels (Fpz-Cz and Pz-Oz).
Since the labeling standard SleepEDF used has been updated, we follow the new standard and merge N4 into N3 \cite{AllanHobson1969ASubjects, Iber2007TheSpecification}. 
We use batch size of 128,  Adam optimizer \cite{Kingma2015Adam:Optimization} and cross-entropy loss.
ISRUC is trained with 100 epochs with learning rate of 4e-4 and SleepEDF uses 40 epochs and learning rate of 2e-4.
For the MC Dropout version, we increase the number of epochs for SleepEDF to 100. 

\textbf{ECG}: 
For the ECG data, we base our model on \cite{Hong2019Mina:Signals}, and replace the last fully connected layer with 4 outputs instead of 2. 
We also added one more Dropout layer with a probability of 0.5 before the fully connected layer before the frequency level attention to handle the overfitting issue.
We also added one more Dropout layer before the final fully connected (output) layer for the same reason.
The model was trained for 100 epochs with the batch size equal to 128, learning rate 3e-3, Adam optimizer, \cite{Kingma2015Adam:Optimization} and cross-entropy loss.
Following the same rationale as in \cite{Hong2019Mina:Signals}, at training time only, rarer classes AF/Others/Noisy are oversampled by 10/2/20x, respectively.

\subsection{Additional Details: Class-specific Risk Experiment}
\textbf{Recording-based Re-sampling}
The convention of processing PSG data is that each recording (or subject) should be completed in training, validation, or test set. 
This is because the patterns tend to be different from recordings to recordings, and training and testing on the same recording will inflate the performance. 
We presented sampling results by epochs in the main text to show what would happen when we have enough data (so that the test and validation sets are not independent but at least identically distributed).
We present the results by splitting data based on the recordings (or subjects, if such information is available)\ref{table:exp_classSpec_ext:split_subjects}.
\TabExpExtendedII
Unfortunately, ISRUC and SleepEDF are relatively small in the number of subjects they have - 16 and 21 left after training, respectively. 
We can see that \methodname still has lower excess risk than LABEL, and the difference usually significant.
However, the key message is that a small number of patients' data might not be enough for class-specific risk control problems for sleep staging tasks. The joint distribution between the base classifier's output on the validation set and the test set is quite different. 
We want to emphasize again that our baselines are \textit{not} designed for this task (except for \methodname-), and as a first step \methodname also has a lot of room to explore and improve.

\textbf{Ambiguities}
It is worth emphasizing that all methods use the same base classifier's output, so we are always facing trade-offs - in this case, we need to sacrifice ambiguity for risks.
Whether the trade-offs are efficient has already been explored in the overall-risk experiment section.
Here, we include the Chance- and Size-ambiguity values in Table \ref{table:exp_classSpec_ext:ambiguities} as a reference to show that although \methodname tends to be more ambiguous, it is never returning degenerate solutions, and ambiguities are generally comparable. %\js{I don't understand last sentence. what do you mean generate answers? Can you rephrase?}
\TabExpExtendedIIAmbiguity

\subsection{Additional Details: Overall Risk Experiment} \label{sec:exp_ext:exp_overall}

\textbf{Synthetic Data}
For each data point, we do the following:
\begin{enumerate}
    \item generate a base random vector $l \sim \mathcal{N}(0,\sigma I_K)$
    \item sample a preliminary class $k'$ uniformly from $[K]$. 
    \item Increase $l_{k'}$ by $\mathbf{s}_{k'}$ where $\mathbf{s}$ is a parameter which can be considered a ``signal strength''. 
    If $\mathbf{s}_{k'}$ is high, it means $k'$ is easy to predict as its conditional probability will usually be closer to 1. 
    \item pass $l$ through Softmax to get a probability distribution $\mathbf{p}$ over the $K$ classes, and sample a label $y$ from the $K$ classes with probability $\mathbf{p}$.
\end{enumerate}
We chose $K=5$ and generated 10,000 data points each for $\validationset$ and $\testset$. 
We set $\mathbf{s}=[9,1,3,3,3]$ (class 0 is easy and class 1 is hard) and $\sigma=3$. 
These parameters are chosen to have a roughly 25\% misclassification risk if we accept the upper bound over the real datasets.

\textbf{Computation of AUC}
Because this paper is not about finding better confidence measures, we will need to control the risk on the $\validationset$ and observe the realized risk on the $\testset$. 
As a result, test risk will be different from validation risk\footnote{e.g., SGR tend to under-realize the risk even when we set its parameter to allow it to exceed risk target 80\% of the time}, and the AUC can only be computed by sampling points on the curve.
\TabExpExtendedIRMSE
To sample the points, we set the risk targets $r$ up to maximum risk (no rejection) with a 1\% stride. %Previously this \% is missing
For example, for ISRUC, we will set the risk target $r$ to be $1\%, 2\%, \ldots, 25\%$. 
For each risk target, we will have a realized (risk, ambiguity) pair. 
We input these sampled points on the curve to \texttt{sklearn.metrics.auc} to compute the AUC.
For the same reason, different methods might have a slightly different span of test risk/ambiguity. 
Since all else equal, AUC is mechanically larger/smaller for larger/smaller ambiguity span methods. To make a fairer comparison, we had to add two anchoring endpoints to all curves to have the same span. 
This might not be necessary if we can evaluate all test risks, which is computationally too expensive.
We also add a third term $\lambda'(\hat{r}(\mathbf{H}) - r^*)^2$ (namely, the overall risk penalty on both directions) with a very small $\lambda'$ that is 1e-4 of $\lambda$. 
This can be seen as imposing a large penalty for excess risk and a tiny penalty for under-realizing risk to sample different points on the curve.

\subsection{Alternative Risk Definitions: Comparison with LABEL}
As suggested in the paper, we can modify the loss a little bit to accommodate different tasks.
In the main paper, we already showed some results using \methodname to control overall risk.
In this section, we will discuss the case of mimicking LABEL. 

LABEL \cite{Sadinle2019LeastLevels} proposes to choose $K$ thresholds to construct $\mathbf{H}$ and solve the following problem:
\begin{problem}\label{eq:opt:label}
\begin{align*}
    \min_{\mathbf{H}} \quad & E[|\mathbf{H}|]\\
    \text{s.t. } \quad & \mathbb{P}_k\{k \not \in \mathbf{H}(X)\} \leq \alpha^*_k, \forall k\in[K]
\end{align*}
\end{problem}
It minimizes the Size-ambiguity while controlling class-specific mis-coverage rates.
The solution is found by using the quantiles on the validation set, like in our method.
Specifically, LABEL chooses $t_k$ to be the $\alpha^*_k$-th quantile among all predicted probability for class $k$ on the validation set:
$$
t_k \defeq \max_{i: y_i = k} \{m_k(x_i): \hat{P}_k\{m_k(X) < m_k(x_i)\}  \leq \alpha^*_k\}
$$

Now, suppose we want to use a similar approach as \methodname for Problem \ref{eq:opt:label}.
If we define the risk for $\mathbf{H}$ (parameterized by $\mathbf{t}$) to be $\alpha_k(\mathbf{H})$, our new loss function will be separable: 
\begin{align}\label{eq:loss:label}
    \hat{L}_{LABEL}(\mathbf{t}) \defeq \underbrace{\hat{E}[|\mathbf{H}|]}_{\text{Size-Ambiguity}} + \sum_{k=1}^K \underbrace{\lambda_k( \hat{\alpha}_k(\mathbf{H}) - \alpha_k^*)_{+}^2}_{\text{class specific mi-coverage penalty}} 
    = \sum_{k=1}^K \hat{L}_{LABEL, k}(t_k)
\end{align}
Where 
$$
\hat{L}_{LABEL, k}(t_k) = \hat{\mathbb{P}}\{m_k(X) > t_k \} + \lambda_k (\hat{\mathbb{P}}_k\{m_k(X) < t_k\} - \alpha_k )_+^2 
$$
is only a function in $t_k$.

If we set $\lambda_k$ to be a large number (for example $>|\validationset|^3$), the mis-coverage rate penalty will need to be satisfied first before the ambiguity comes into play, and the full optimization will only perform $K$ descent iterations to find the global optimum.
Obviously, for this particular problem, it will be more efficient to select the quantiles according to \cite{Sadinle2019LeastLevels}. 
However, we include this section as to another example of why the underlying method of \methodname is flexible.
It is also clear that the LABEL paper is more about the optimality of this particular problem: LABEL provides an intuitive and simple justification for using the particular quantiles, but it does not provide flexible tools to tackle anything different. 
In particular, the problem \textit{must} be separable into multiple one-versus-all binary classification problems like we did in rewriting $\hat{L}_{LABEL, k}$.

\subsection{Chance-Ambiguity vs. Size-Ambiguity} 
In this experiment, we randomly split the data not used in training into validation and test sets. 
Then, we randomly pick $K$ thresholds on the validation set and evaluate the ambiguities of the corresponding $\mathbf{H}$ on the test set.
We compute the Pearson and Spearman correlations between Chance- and Size-Ambiguities on 1,000 $\mathbf{H}$ generated this way. 
We then repeat the experiments 20 times for mean and standard deviations. 
As shown in Table \ref{table:exp_ext:ambireal}, correlations are generally between 80\% and 90\% for the datasets we have.
\AmbiguityCorrelation

\subsection{Comparison to Other Optimization Methods} \label{appendix:subsec:exp:optcompare}
\TabOptimizationCompareNew
\FigOptimizationCompare
In addition to the coordinate descent method we currently have, we also tried the following optimization methods:
\begin{itemize}
    \item Bayesian Optimization with Gaussian processes (implemented by \cite{BayesianOptimizationLib})
    \item Truncated Newton (TNC) \cite{Nash1984NEWTON-TYPEMETHOD.}
    \item Modified Powell (Powell) \cite{Powell1964AnDerivatives}
    \item Limited-memory BFGS (L-BFGS-B) \cite{Byrd1995AOptimization}
\end{itemize}
Moreover, after finding the local minimum $\mathbf{t}^*$, our method currently sample 1000 random $\mathbf{t}$ within $\pm10\%$ of the quantiles
\footnote{For example, if $N=1000$ and $t^*_1 = \mathbf{M}_{300, 1}$ where $\mathbf{M}$ means the same as in Algorithm \ref{alg:search_each_iter}, then in the sampling $t_1$ will be drawn from $\{\mathbf{M}_{j, 1}\}_{j=200}^{400}$}.
This improves the final result but adds to the computation time. 
We include a variant of our method that removes this sampling step, called ``Ours-'', in the comparison. 

TNC, Powell, L-BFGS-B are the only methods in the \texttt{scipy} library that support bounds, and we used the implementation in \texttt{scipy} \cite{2020SciPy}.

Like in the coordinate descent, we repeat the optimization several times and pick the lowest loss.
For Bayesian optimization, we use 10 initial points and 100 iterations.
For all other methods (including ours), we use the best from 10 random initial points in each optimization.

All parameters for TNC/Powell/L-BFGS-B are the default in \texttt{scipy.optimize.minimize}. 

The comparison is in Table \ref{table:exp_opt:loss_and_time_new}.
All experiments were carried out on an Intel Broadwell CPU.
Among all methods, only Powell is sometimes close to ours in terms of the loss value it founds. 
The rest of the methods tend to return high final loss.
In terms of time spent, ours is not the best, but comparable with Powell and much faster than Bayesian Optimization.
Since optimization is not a bottleneck in our experiments (at most a few seconds), we conclude that for our task, our optimization method finds the best optima with reasonable time.

\bibliography{references,ref1}
\bibliographystyle{apalike}

\end{document}